\documentclass[runningheads]{llncs}

\usepackage{eccv}

\usepackage{eccvabbrv}
\usepackage{multirow}
\usepackage{graphicx}
\usepackage{booktabs}

\usepackage[accsupp]{axessibility}  % Improves PDF readability for those with disabilities.

\def\our{GaINeR}
\def\fullname{Geometry-Aware Implicit Network Representation}

\usepackage{array}
\usepackage{adjustbox}
\newcolumntype{R}[2]{%
    >{\adjustbox{angle=#1,lap=\width-(#2)}\bgroup}%
    l%
    <{\egroup}%
}
\usepackage{pifont}% http://ctan.org/pkg/pifont

\def\x{\mathbf{x}}

\def\G{\mathcal{G}}

\usepackage{hyperref}

\usepackage{orcidlink}

\begin{document}

\title{\our{}: Geometry-Aware Implicit Neural Representation for Image Editing} 

\titlerunning{Geometry-Aware Implicit Neural Representation for Image Editing}

\author{
Weronika Jakubowska\inst{1}$^*$\and
Mikołaj Zieliński\inst{2}$^*$\and
Rafał Tobiasz\inst{3,4}$^*$\and
Krzysztof Byrski\inst{4}\and
Maciej Zięba\inst{1}\and
Dominik Belter\inst{2}\and
Przemysław Spurek\inst{3,4}
}
\authorrunning{Jakubowska et al.}

\institute{
Wrocław University of Science and Technology \and
Poznań University of Technology \and
IDEAS Research Institute \and
Jagiellonian University\\[6pt]
$^*$Equal contribution\\
\email{weronika.jakubowska@pwr.edu.pl}
}

\maketitle

\begin{center}
    \captionsetup{type=figure}
    \vspace{-0.5cm}
    \includegraphics[width=0.9\textwidth]{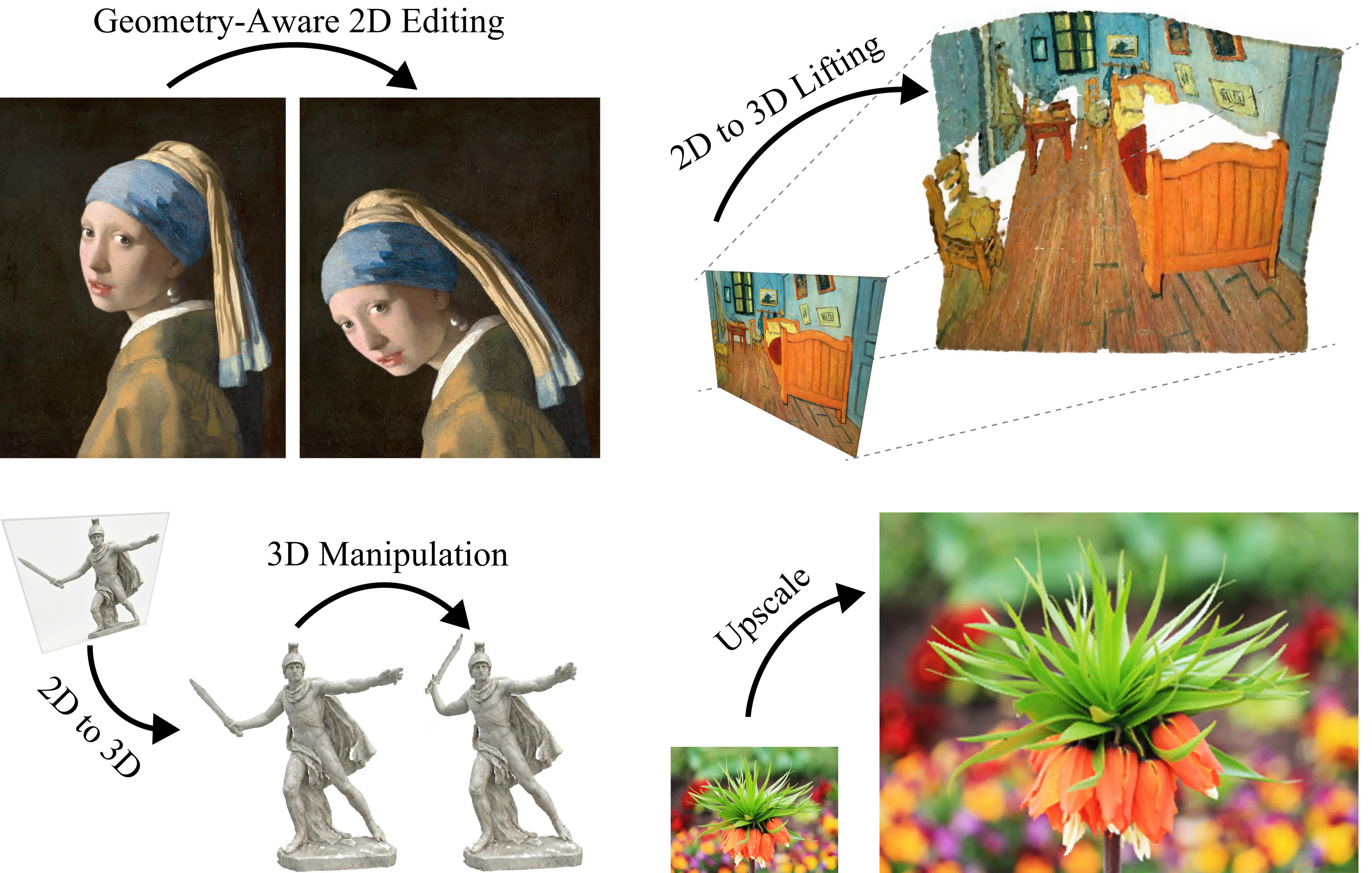}
    \caption{
    Given a single image, \our{} reconstructs a manipulable representation that enables geometry-aware editing. Our method supports object-level edits (top left), 2D to 3D transformations (top right), editing 3D reconstruction (bottom left), and upscaling (bottom right) while preserving semantic structure and photometric coherence.
    }
    \label{fig:teaser}
\end{center}

\begin{abstract}
  % Implicit Neural Representations (INRs) have become an essential tool for modeling continuous 2D images, enabling high-fidelity reconstruction, super-resolution, and compression. Popular architectures such as SIREN, WIRE, and FINER demonstrate the potential of INR for capturing fine-grained image details. However, traditional INRs often lack explicit geometric structure and have limited capabilities for local editing, 2D-to-3D transformations, and integration with physical simulation, thereby restricting their applicability in dynamic or interactive settings. To address these limitations, we propose \our{} (\fullname{}), a novel framework for 2D images that combines trainable Gaussian distributions with a neural network-based INR. For a given image coordinate, the model retrieves the K nearest Gaussians, aggregates distance-weighted embeddings, and predicts the RGB value via a neural network. This design enables continuous image representation, interpretable geometric structure, and flexible local editing, providing a foundation for physically aware and interactive image manipulation.

Implicit Neural Representations (INRs) are widely used for modeling continuous 2D images, enabling high-fidelity reconstruction, super-resolution, and compression. Architectures such as SIREN, WIRE, and FINER demonstrate their ability to capture fine image details. However, conventional INRs lack explicit geometric structure, limiting local editing, and integration with physical simulation. To address these limitations, we propose \our{} (\fullname{}), a novel framework for 2D images that combines trainable Gaussian distributions with a neural network-based INR. For a given image coordinate, the model retrieves the K nearest Gaussians, aggregates distance-weighted embeddings, and predicts the RGB value via a neural network.  This design enables continuous image representation, interpretable geometric structure, and flexible local editing, providing a foundation for physically aware and interactive image manipulation. Our method supports geometry-consistent transformations, seamless super-resolution, and integration with physics-based simulations. Moreover, the Gaussian representation allows lifting a single 2D image into a geometry-aware 3D representation, enabling depth-guided editing. Experiments demonstrate that \our{} achieves state-of-the-art reconstruction quality while maintaining flexible and physically consistent image editing. The official implementation of our method is publicly available at \url{https://github.com/WJakubowska/GaINeR}.
  
\end{abstract}

%%%%%%%%%%%%%%%%%%%%%%%%%%%%%%%
\section{Introduction}
%%%%%%%%%%%%%%%%%%%%%%%%%%%%%%%

% Implicit Neural Representations, have fundamentally reshaped how continuous signals are represented and processed across domains ranging from computer vision to computational physics \cite{klocek2019hypernetwork,sitzmann2020implicit,takikawa2023neural}. Unlike traditional discrete representations tied to fixed spatial resolutions, INRs parameterize signals as continuous functions via neural networks, providing resolution-independent representations with high expressiveness and memory efficiency. 
% Implicit Neural Representations (INRs) have transformed the modeling of continuous signals across domains such as computer vision and computational physics \cite{klocek2019hypernetwork,sitzmann2020implicit,takikawa2023neural}. Instead of discrete grids tied to fixed spatial resolutions, INRs represent signals as continuous functions parameterized by neural networks, enabling resolution-independent representations with high expressiveness and compact memory usage.
% While several architectures, such as SIREN \cite{sitzmann2020implicit}, GAUSS \cite{ramasinghe2022beyond}, WIRE \cite{saragadam2023wire}, FINER \cite{liu2024finer}, and positional encoding \cite{tancik2020fourier} have improved the ability of INRs to capture high-frequency details in images, they typically lack explicit geometric structure, limiting the possibility of local editing and integration with physical simulations.
Implicit Neural Representations (INRs) have transformed the modeling of continuous signals across domains such as computer vision and computational physics \cite{klocek2019hypernetwork,sitzmann2020implicit,takikawa2023neural}. Instead of discrete grids tied to fixed spatial resolutions, INRs represent signals as continuous functions parameterized by neural networks, enabling resolution-independent representations with high expressiveness and compact memory usage. While architectures such as SIREN \cite{sitzmann2020implicit}, GAUSS \cite{ramasinghe2022beyond}, WIRE \cite{saragadam2023wire}, FINER \cite{liu2024finer}, and positional encoding \cite{tancik2020fourier} improve the ability of INRs to capture high-frequency image details, they typically lack explicit geometric structure, limiting local editing and integration with physical simulations.

Recent work, MiRaGe \cite{waczynskamirage}, addresses image editability by representing 2D images by a Gaussian Splatting-based model, enabling intuitive modifications and coupling with physics engines. Each Gaussian acts as a discrete component, allowing local transformations and perspective changes. However, because the representation is discrete, MiRaGe suffers from producing artifacts when the physical engine significantly alters the geometry of objects in a 2D image. The problem can be partially solved by training very small Gaussians, but it requires dedicated parameters that must be manually tuned. Additionally a Gaussian Splatting-based model cannot provide a truly continuous image representation.

\begin{figure}[t]
\centering
    \includegraphics[width=\linewidth]{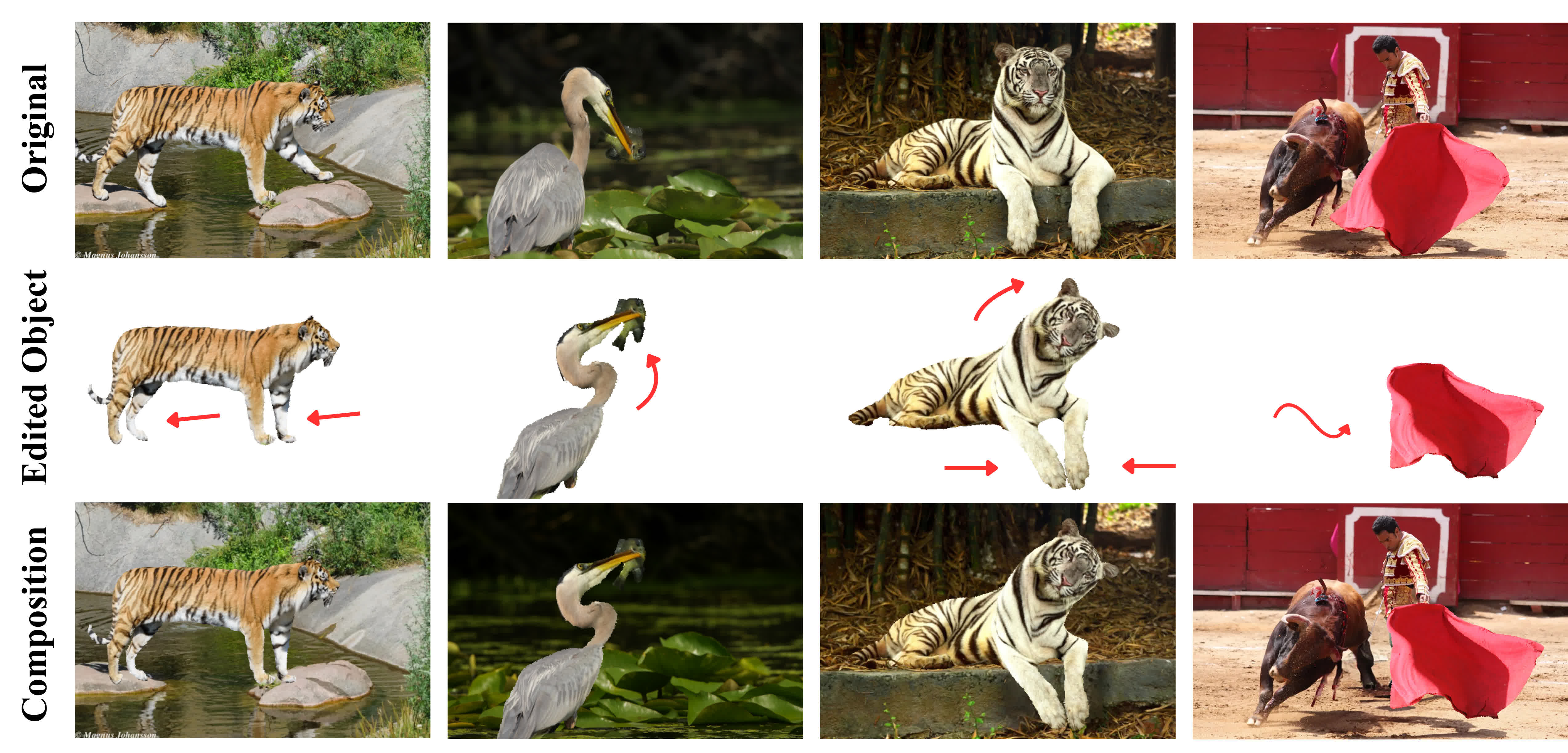}
    \caption{
    \our{} produces realistic image edits by preserving structural consistency under spatial transformations. Red arrows indicate the direction of the applied changes.
    }
    \label{fig:editing_examples}
    \vspace{-0.3cm}
\end{figure}

To overcome these limitations, we propose \fullname{} (\our{}), the first model that combines the continuous representation of INR for 2D images with trainable Gaussian embeddings. \our{} enables editing 2D images, and 2D to 3D manipulations (see Fig.~\ref{fig:teaser}, Fig.~\ref{fig:editing_examples}). We achieve this by explicitly conditioning the implicit function on a geometric structure defined by a set of specific trainable Gaussian components, each parameterized by position, scale, and a unique learnable feature embedding. The core INR network then acts as a decoder. Given a query coordinate $(x, y)$, the network retrieves and aggregates the features of the $k$ nearest Gaussians using a distance-weighted mechanism (see Fig.~\ref{fig:model_overview}). The MLP then uses this aggregated geometric and feature information to synthesize the final, high-fidelity RGB colors. This design allows precise spatial manipulation of the image, as Gaussian positions and shapes can be adjusted to produce smooth, real-time deformations. Additionally, the Gaussians can act as discrete physical entities, enabling the entire 2D representation to interact naturally with standard physics engines for dynamic simulation and animation. As a result, \our{} achieves state-of-the-art performance in 2D image reconstruction, while maintaining fully continuous and editable representations.
Importantly, due to the locality of Gaussian conditioning, our representation naturally extends beyond the 2D image plane. By treating the image as a continuous surface embedded in 3D space, \our{} enables lifting a single 2D image into a geometry-aware 3D representation. This allows depth-guided Gaussian displacement and 3D-consistent editing. Furthermore, as a fully continuous representation, \our{} inherently supports arbitrary-resolution image synthesis, enabling seamless super-resolution and upscaling without retraining.
% A visual overview of our framework and its editing mechanism is shown in Fig.~\ref{fig:teaser}.

The following constitutes a list of our key contributions:
\begin{itemize}
    \item We introduce \our{}, an implicit neural representation of 2D images based on Gaussian embeddings and MLPs, which provides superior reconstruction quality compared to prior INR-based approaches.
    % \item We propose using Gaussians as feature providers conditioning an INR decoder, enabling high-fidelity and continuous color reconstruction.
    \item We extend the INR framework by introducing a geometric component that allows direct and flexible 2D image modification and integrates with a physical simulation engine for interactive effects.  Thanks to its fully continuous formulation, \our{} also supports arbitrary-resolution image synthesis, enabling seamless super-resolution and upscaling without retraining.
    \item We demonstrate that our geometry-aware formulation enables single-image 3D-aware transformations via depth-guided Gaussian displacement, allowing consistent spatial manipulation beyond the 2D image plane.
    
    % \item Thanks to its fully continuous formulation, \our{} supports arbitrary-resolution 2D image synthesis, enabling seamless super-resolution and upscaling without retraining.

    %    \item We introduce \our{}, an implicit neural representation of 2D images that leverages Gaussian embeddings as feature providers conditioning an MLP decoder, enabling high-fidelity and fully continuous image reconstruction that outperforms prior INR-based approaches.

    % \item We extend the INR framework with a geometry-aware formulation that enables direct and flexible 2D image manipulation by operating on Gaussian primitives, while also integrating with a physical simulation engine for interactive effects.

    % \item We demonstrate that the proposed representation supports consistent spatial editing, including single-image 3D-aware transformations via depth-guided Gaussian displacement, and arbitrary-resolution image synthesis such as seamless super-resolution and upscaling without retraining.
\end{itemize}

%%%%%%%%%%%%%%%%%%%%%%%%%%%%%%%
\section{Related work}
%%%%%%%%%%%%%%%%%%%%%%%%%%%%%%%

Implicit Neural Representations (INRs) have received considerable attention in research \cite{tewari2022advances} as neural models designed for signal representation. This paradigm has found broad application across various domains, including the representation of images \cite{klocek2019hypernetwork}, videos \cite{chen2022videoinr}, and $3D$ shapes \cite{park2019deepsdf}. Significant uses span $3D$ shape reconstruction \cite{mildenhall2021nerf}, robotics \cite{lin2021barf}, and data compression \cite{takikawa2023neural}. 

\textbf{INR for images}
INR architectures often remain simple, typically consisting of a single Multi-Layer Perceptron (MLP), with recent efforts focused on enhancing capacity through specialized embedding techniques \cite{chen2022tensorf,muller2022instant}, unique activation functions \cite{sitzmann2020metasdf}, refined rendering methods~\cite{barron2021mip}, and regularization strategies~\cite{yang2023freenerf}.

Implicit Neural Representations (INRs), or coordinate-based MLPs, have fundamentally transformed the representation and processing of continuous signals across various domains. 
The main direction of INRs development for 2D images is to adapt various activation functions and training procedures. 
SIREN \cite{sitzmann2020implicit} uses sine activations alongside a frequency hyperparameter $\omega$, enabling smooth fitting and stable training. While effective for continuous signals, its fixed-frequency multiplier limits versatility, resulting in lower accuracy when modeling details across diverse frequencies \cite{kaniafresh}. GAUSS \cite{ramasinghe2022beyond} employs a Gaussian activation function providing smooth localized representations that are beneficial for signal reconstruction and denoising. However, its aperiodic nature hinders the effective modeling of sharp, high-frequency details. WIRE \cite{saragadam2023wire} utilizes the Gabor wavelet activation, combining sinusoidal and Gaussian components to capture localized frequencies. Although effective for modeling high-frequency details within specific spatial regions, its complex-valued formulation increases computational demands, slowing both training and inference. FINER \cite{liu2024finer} modifies SIREN with a frequency-variable activation, broadening the representable frequency spectrum while retaining SIREN's benefits. However, FINER is sensitive to bias initialization and complicates the training process. Lastly, Positional Encoding (PE) \cite{tancik2020fourier} maps inputs into a higher-dimensional space using sinusoidal functions to mitigate the spectral bias common in MLPs. 

Such a representation provides high-quality reconstruction but offers limited manual editing and integration with physical engines. In this paper, we aim to incorporate a geometric component into the INR framework to facilitate 2D image modification. 

\textbf{Explicit and Editable Representations} 
Although INRs excel at compression and continuous representation, their inherent implicit nature severely limits direct interaction and editing. Conversely, explicit representations prioritize geometry and manipulability. $3D$ Gaussian Splatting (3DGS)~\cite{kerbl3Dgaussians} recently revolutionized novel view synthesis by employing a vast collection of explicit, $3D$ anisotropic Gaussians, achieving real-time rendering speeds. This approach naturally allows direct manipulation of Gaussian primitives, offering superior editability compared to traditional volumetric methods. Following this, GaussianImage~\cite{zhang2024gaussianimage} adapted the core GS concept to a $2D$ image representation, demonstrating competitive compression and quality. However, while these Gaussian-based methods introduce necessary explicit geometry, they still rely on simple color and opacity stored directly within each primitive. Consequently, they lose the continuous spatial coherence and superior color fidelity offered by a dedicated INR decoder. Models such as MiRaGe~\cite{waczynskamirage} attempted to address the editability challenge for $2D$ images by placing $2D$ Gaussian representations in a $3D$ space to enable physical interactions such as bending and folding. 

Our \our{} framework distinguishes itself from these methods by creating a deeper synergy, where the explicit geometric primitives (Gaussians) are not renderers themselves, but rather rich feature providers that condition a dedicated Implicit Neural Representation decoder, thus ensuring both high-fidelity continuous color and native geometric editability and physical readiness. Examples of realistic edits enabled by our method are illustrated in Fig.~\ref{fig:editing_examples}.

%%%%%%%%%%%%%%%%%%%%%%%%%%%%%%%
\section{Method}
%%%%%%%%%%%%%%%%%%%%%%%%%%%%%%%

\begin{figure*}[t]
\centering
    \includegraphics[width=\linewidth]{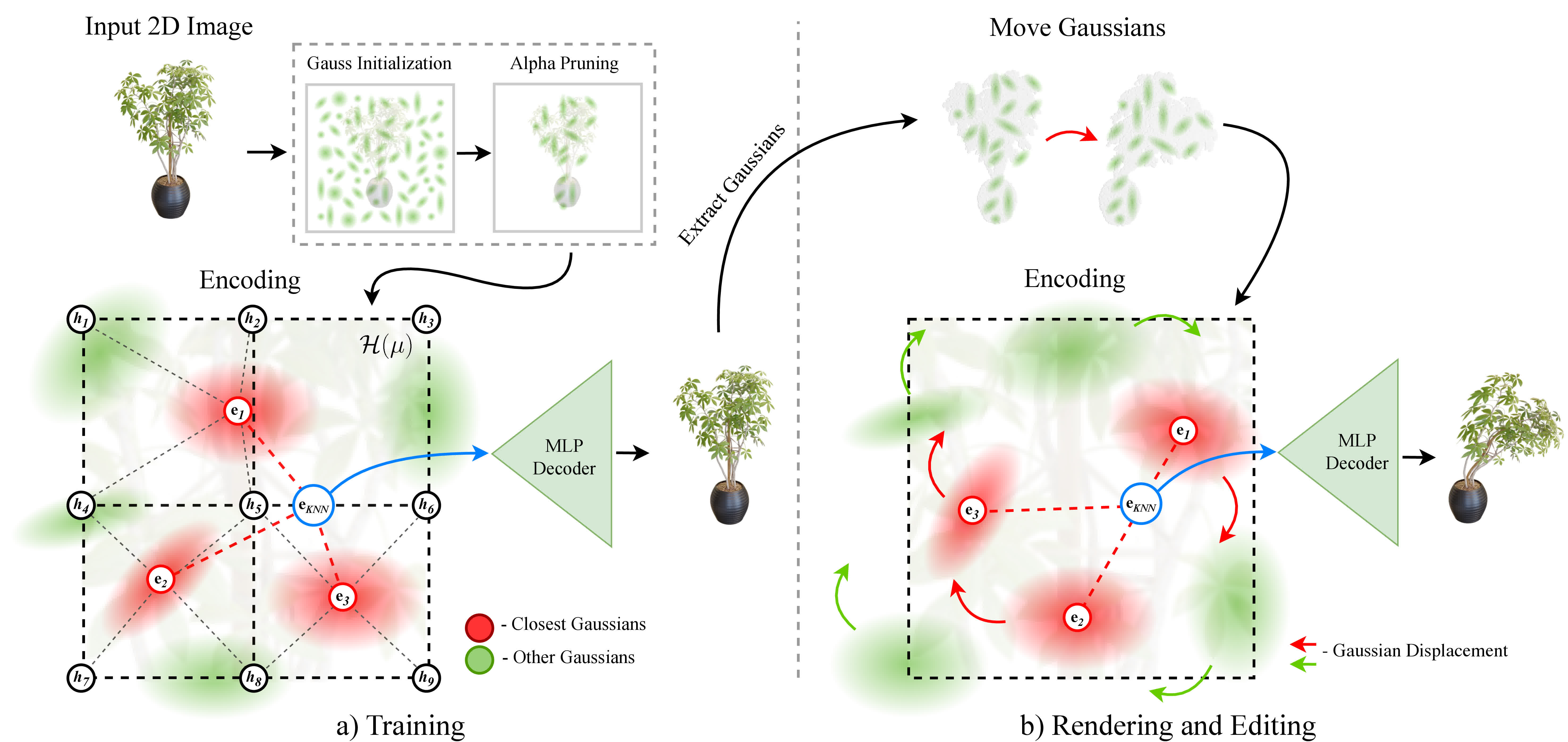}
    \caption{
    Overview of the \our{} framework.  
    (a) \textbf{Training:} For RGBA inputs, Gaussian components outside the alpha mask are removed from training.  
    Each input coordinate $\mathbf{x}$ is encoded via Gaussian features $\mathbf{e}_i$ derived from the multi-resolution hashgrid $\mathcal{H}(\mu_i)$.  
    The aggregated embedding $\mathbf{e}_{\mathrm{KNN}}(\mathbf{x}, \mathcal{G})$ is obtained through radius-limited KNN interpolation and decoded by the MLP $f_\theta$ to reconstruct pixel color $\mathbf{c}$.  
    (b) \textbf{Rendering and Editing:} After training, the optimized Gaussian set $\mathcal{G}$ is used to reconstruct the sampling mask (for RGBA cases) and to enable spatial transformations.  
    Edited Gaussian means $\mu_i' \in \mathbb{R}^2$ are re-encoded to produce updated embeddings $\mathbf{e}_{\mathrm{KNN}}'(\mathbf{x}, \mathcal{G}')$, which the decoder maps to modified image values, enabling geometry-consistent editing.
    }
    \vspace{-0.2cm}
    \label{fig:model_overview}
\end{figure*}

In this section, we present the \our{} framework (see Fig. \ref{fig:model_overview}).  
We begin by revisiting the classical formulation of Implicit Neural Representations (INRs) to establish the foundational concepts.  
Next, we introduce \our{}, which incorporates learnable Gaussian embeddings to achieve spatially consistent and editable 2D representations.  
Finally, we describe the training objective and optimization strategy used to jointly learn the embedding distribution and neural decoder.

While \our{} is introduced in the context of 2D image reconstruction and editing, its formulation is inherently dimension-agnostic.
Since Gaussian embeddings and KNN-based aggregation operate in continuous space, the same mechanism naturally extends to three-dimensional domains \cite{muller2022instant}.
Consequently, \our{} generalizes seamlessly from 2D image representations to 3D volumetric modeling.

\textbf{Classical Implicit Neural Representations}
Implicit Neural Representations provide a continuous and compact way to encode signals such as images, shapes, or radiance fields using neural networks. Instead of storing discrete samples (e.g., pixels or voxels), an INR models the underlying signal as a function \( f_\theta \) that directly maps input coordinates to output values. This formulation enables subpixel-level precision, smooth interpolation, and differentiability with respect to spatial coordinates.

Formally, an INR is typically realized as a multilayer perceptron (MLP):
\[
f_\theta : \mathbb{R}^d \rightarrow \mathbb{R}^m,
\]
where \( d \) denotes the dimensionality of the input domain (e.g., spatial coordinates), and \( m \) corresponds to the output space (e.g., color channels or density).  
For 2D images, the INR learns a mapping
\[
f_\theta(x, y) = (r, g, b),
\]
which associates each coordinate \((x, y)\) with an RGB color value.  
In Neural Radiance Fields (NeRFs), the input extends to a 3D position and a viewing direction:
\[
f_\theta(\x, \mathbf{d}) = (c, \sigma),
\]
where \( c \) represents the emitted color and \( \sigma \) is the corresponding volume density.

% Following the standard MLP formulation, an INR can be expressed as a composition of layer functions:
% \[
% f_\theta(\x) = f_{\theta_L} \circ \cdots \circ f_{\theta_1}(\x),
% \]
% where each layer function \( f_{\theta_i} : \mathbb{R}^{n_{i-1}} \rightarrow \mathbb{R}^{n_i} \) is parameterized by the subset of weights \( \theta_i \). The hidden activations are recursively defined as:
% \[
% h_i = f_{\theta_i}(h_{i-1}) = \sigma(a_{\theta_i}(h_{i-1})), \quad i = 1, \ldots, L-1,
% \]
% where \( a_{\theta_i} : \mathbb{R}^{n_{i-1}} \rightarrow \mathbb{R}^{n_i} \) denotes an affine transformation, and \( \sigma \) is a nonlinear activation function such as sine, ReLU, or softplus.  
% The final layer is typically linear:
% \[
% f_{\theta_L}(h_{L-1}) = a_{\theta_L}(h_{L-1}).
% \]

% \begin{figure}[t]
% \centering
%     \includegraphics[width=\linewidth]{figures/plot_psnr.png}
%     \caption{
%         Comparison of PSNR obtained on a Singapore image from
%         DIV2K dataset. All models were trained for 30 k iterations, yet our method rapidly achieves the highest PSNR within the first 1–2 k iterations, often already surpassing all other approaches.
%     }
%     \label{fig:plot_psnr}
% \end{figure}

This purely functional formulation captures all spatial information implicitly within the network parameters \( \theta \), without requiring discrete grids or explicit coordinate encoding.  
Classical INRs exhibit several key limitations. First, they lack explicit geometric structure, making it difficult to reason about spatial relationships or integrate with physical simulation engines. Second, the global nature of MLP parameterization restricts local controllability, complicating fine-grained image editing or deformation. Finally, reconstruction quality often depends on careful architectural tuning and positional encodings, which limit generalization across diverse image domains.  
These challenges motivate the development of geometry-aware extensions, such as our proposed \our{} model, which bridges implicit representations with explicit geometric priors and physical consistency.

\textbf{\fullname{}}
Our method extends the classical INR formulation by introducing a geometry-consistent embedding mechanism based on a family of learnable Gaussian distributions.  
Each 2D coordinate in the image domain is represented not as a fixed input vector, but as a continuous Gaussian embedding that encodes its local spatial context.  
This design enables geometry-aware reasoning and smooth interpolation while preserving the functional nature of implicit representations.

Let the input domain be \(\x = (x, y) \in \mathbb{R}^2\).  
We define a set of \(N\) Gaussian components, each parameterized by its mean \(\mu_i \in \mathbb{R}^2\), covariance matrix \(\Sigma_i \in \mathbb{R}^{2 \times 2}\), and a learnable embedding vector \(\mathbf{e}_i \in \mathbb{R}^d\):
\[
\G = \{ (\mathcal{N}_i (\mu_i, \Sigma_i), \mathbf{e}_i) \}_{i=1}^{N}.
\]

To enhance the representational capacity of the embeddings, we are using multi-resolution hashgrid features \cite{muller2022instant} as Gaussian features obtained from,  
$\mathcal{H}(\mu) = \mathbf{e},$
where $\mathcal{H}$ is hashing function. 
These features provide fine-grained spatial detail during training but are omitted at inference, as the learned Gaussian parameters themselves encode the required multi-scale structure.

For a given input coordinate \(\x\), we identify its \(K\) nearest Gaussian centers in Euclidean space using a radius-limited \(k\)-nearest neighbors (KNN) search, where candidates are considered only within a distance \(r\) for computational efficiency.  
The corresponding embedding vectors of these neighbors are then aggregated using Gaussian-weighted averaging:

\[
\mathbf{e}_{\mathbf{KNN}}(\x, \G) = 
\frac{
\sum_{i \in \mathbf{KNN}(\x, \G)} 
\mathcal{N}_i(\mathbf{x};\mu_i, \Sigma_i, \mathbf{e}_i) \, \mathbf{e}_i
}{
\sum_{i \in \mathbf{KNN}(\x, \G)} 
\mathcal{N}_i(\mathbf{x};\mu_i, \Sigma_i, \mathbf{e}_i)
},
\]
where \(\mathbf{KNN}(\x, \G)\) denotes the indices of the \(K\) nearest neighbors of \(\x\).  
The resulting feature
\[
\mathbf{e}(\x) = \mathbf{e}_{\mathbf{KNN}}(\x, \G) \in \mathbb{R}^d
\]
serves as a locally geometry-aware embedding that captures both spatial proximity and the density structure of the learned Gaussians.

The aggregated embedding \(\mathbf{e}(\x)\) is subsequently processed by a multilayer perceptron (MLP) decoder:
\[
f_\theta(\x) = \mathrm{MLP}_\theta(\mathbf{e}(\x)),
\]
where \(f_\theta : \mathbb{R}^d \rightarrow \mathbb{R}^3\) produces the RGB color value corresponding to the coordinate \(\x\).  
This combination of continuous Gaussian embeddings and coordinate-based neural decoding allows the model to reconstruct and edit 2D images with smooth, geometry-consistent variations.

Unlike classical INRs, which operate directly on raw spatial coordinates, \our{} embeds each position into a learned, spatially coherent feature space.  
This design enhances the model’s ability to capture local geometric relationships, improves reconstruction fidelity, and facilitates controllable 2D image editing through interpretable, geometry-aware embeddings.

\begin{figure}[]
\centering
    \includegraphics[width=\linewidth]{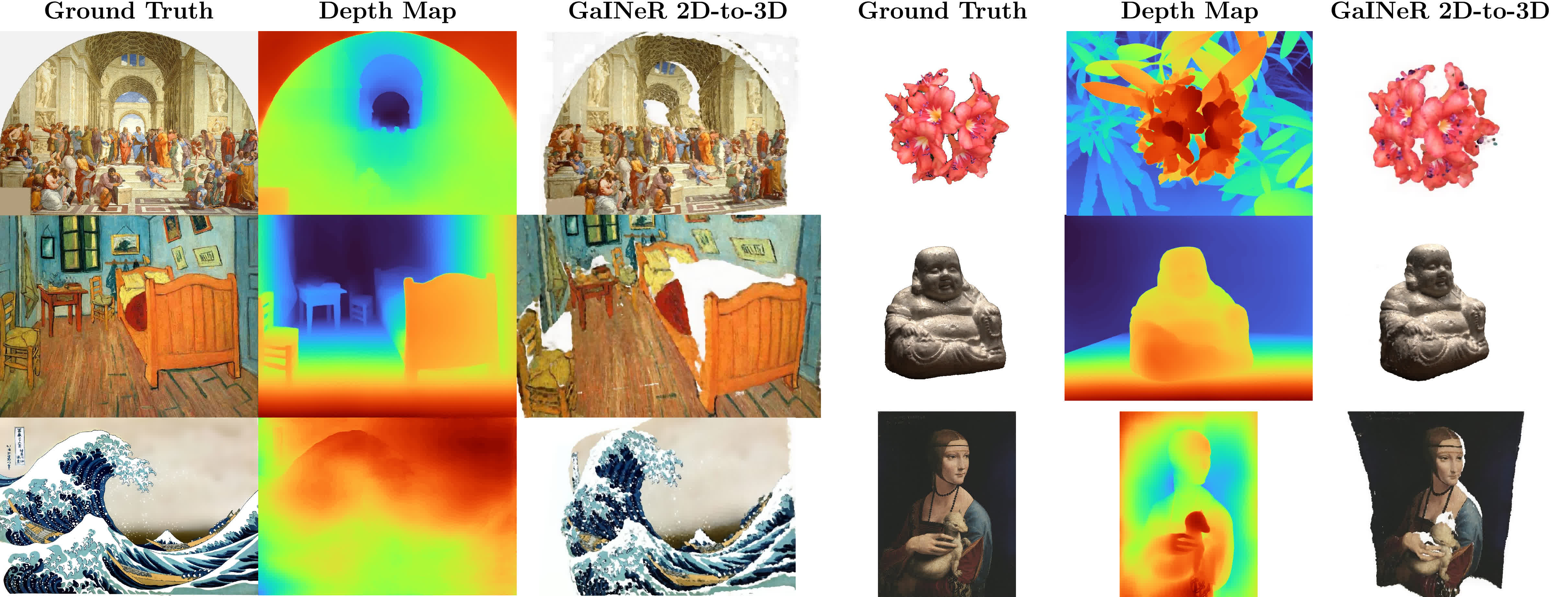}
    \caption{
        Results of 2D-to-3D lifting with \our{}. We elevate flat 2D images into fully volumetric 3D representations using monocular depth maps. The final column shows the reconstructed geometry rendered from a novel viewpoint. Our method generalizes well across complex paintings and masked objects.
    }
    \vspace{-0.3cm}
    \label{fig:2d_to_3d}
\end{figure}

\textbf{Training Objective}
The training of \our{} jointly optimizes the parameters of the Gaussian embedding 
$\{(\Sigma_i, \mathbf{e}_i) \}_{i=1}^N$,
the multi-resolution hashgrid encoder \(\mathcal{H}(\cdot)\),
and the MLP decoder \(f_\theta\) that reconstructs the target image.

Given a training set $\{ (\mathbf{x}_j, \mathbf{c}_j) \}_{j=1}^{M}$, where $M \in N$ is a subset of Gaussians in training dataset, $\mathbf{x}_j = (x_j, y_j)$ denotes pixel coordinates, and $\mathbf{c}_j = (r_j, g_j, b_j)$ are the corresponding RGB values, the model is trained to minimize the Smooth L1 reconstruction loss:
\[
\mathcal{L}_{\mathrm{rec}} =
\frac{1}{M}
\sum_{j=1}^{M}
\operatorname{SmoothL1}\!\left(
f_\theta(\mathbf{x}_j) - \mathbf{c}_j
\right),
\]
where
\[
\operatorname{SmoothL1}(\mathbf{r}) =
\begin{cases}
\frac{1}{2} \|\mathbf{r}\|_2^2 / \beta, & \text{if } \|\mathbf{r}\|_1 < \beta, \\
\|\mathbf{r}\|_1 - \frac{1}{2}\beta, & \text{otherwise},
\end{cases}
\]
and $\mathbf{r} = f_\theta(\mathbf{x}_j) - \mathbf{c}_j$ denotes the residual, while $\beta$ controls the transition between the $\ell_2$ and $\ell_1$ regions.

% To ensure numerical stability and efficient optimization, we decompose each covariance matrix into its eigenbasis:
% \[
% \Sigma_i = R_i S_i R_i^\top,
% \]
% where \(R_i \in \mathbb{R}^{2 \times 2}\) is a rotation matrix and \(S_i = \mathrm{diag}(s_{i,1}^2, s_{i,2}^2)\) contains the eigenvalues representing the principal variances.  
% This parameterization guarantees positive semi-definiteness and allows for anisotropic, spatially oriented embeddings that adapt to local image structures.

\textbf{Alpha-based Sampling and Mask Reconstruction.}
For images containing an alpha channel, i.e., RGBA inputs, we utilize the alpha map as a spatial sampling mask during training to improve computational efficiency.  
Specifically, only pixels with nonzero alpha values are included in the training set 
$\{ (\mathbf{x}_j, \mathbf{c}_j) \}_{j=1}^{M}$,
which focuses optimization on visible regions and avoids redundant computation in transparent areas.

During inference, the corresponding mask must also be reconstructed to enable consistent rendering and editing.  
To this end, we also train a dedicated mask model that learns to reproduce the alpha-based sampling mask.  
This model is subsequently animated in synchrony with the image model, ensuring temporally coherent and spatially accurate masks for all generated or edited frames.
    
\textbf{Image Editing}
During inference, the hashgrid encoder \(\mathcal{H}\) is no longer required, since the learned Gaussian embeddings \(\mathbf{e}_i\) already encode multi-scale spatial information acquired during training.  
Thus, the model operates directly on the optimized Gaussian parameters
\(\G = \{ \mathcal{N}(\mu_i, \Sigma_i, \mathbf{e}_i) \}_{i=1}^{N}\),
enabling efficient and interpretable scene editing.

Image editing in \our{} is performed by modifying the spatial configuration of the Gaussian means \(\mu_i\).  
Because the feature associated with any query coordinate \(\x\) is computed through a radius-limited KNN aggregation, \(\mathbf{e}_{\mathbf{KNN}}(\x, \G),\)
shifting a subset of Gaussian centers \(\mu_i \mapsto \mu_i'\) effectively alters the local neighborhood structure and, consequently, the interpolated feature \(\mathbf{e}_{\mathbf{KNN}}(\x, \G)\).  
The MLP decoder \(f_\theta\) then produces a modified color field corresponding to the updated embedding distribution:
\[
f_\theta'(\x) = \mathrm{MLP}_\theta(\mathbf{e}_{\mathbf{KNN}}(\x, \G')).
\]

This formulation allows geometric transformations such as translation, rotation, bending, or stretching to be applied directly in the Gaussian space while maintaining photometric consistency in the output.  
Since the MLP decoder is conditioned only on the aggregated embeddings, not on absolute Gaussian positions, the model exhibits a form of spatial equivariance: moving the Gaussians induces coherent deformations in the reconstructed image without introducing artifacts.

\begin{figure}[h]
\centering
    \includegraphics[width=\linewidth]{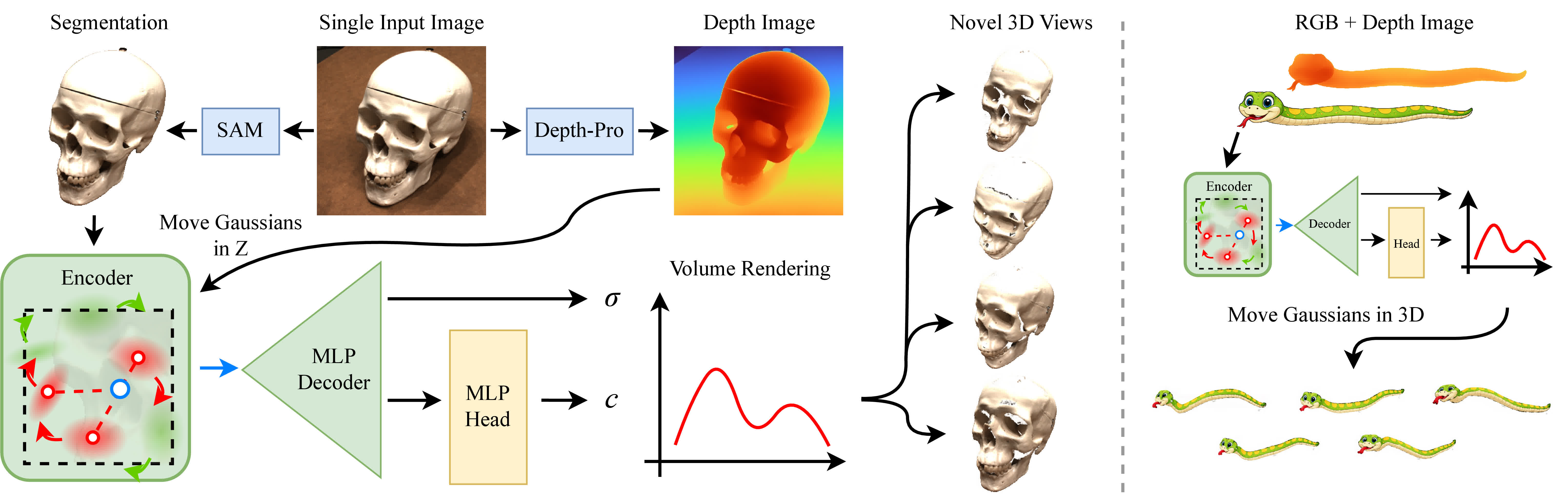}
    \caption{
    % Overview of 3D reconstruction and rendering with \our{}.  
    % Input images are first segmented using Segment Anything (SAM)~\cite{kirillov2023segment} and depth is estimated with Depth-Pro~\cite{bochkovskiy2025depthpro}.  
    % The model is trained in 2D, but during inference the Gaussian means $\mu_i$ are lifted into 3D using the estimated depth values.  
    % The decoder is extended to predict both color $\mathbf{c}$ and density $\sigma$, enabling volumetric rendering analogous to NeRF~\cite{mildenhall2021nerf}.  
    % This allows the model to synthesize consistent 3D views from novel camera perspectives, as illustrated at the bottom of the figure.
    Overview of 3D reconstruction with \our{}. Input images are segmented with SAM~\cite{kirillov2023segment} and lifted to 3D using depth estimated by Depth-Pro~\cite{bochkovskiy2025depthpro}. The Gaussian means are projected into 3D while the decoder predicts color $c$ and density $\sigma$, enabling NeRF-style volumetric rendering and novel-view synthesis, and direct 3D scene modifications.
    }
    \label{fig:model_3d}
    \vspace{-0.3cm}
\end{figure}

\textbf{2D To 3D Transformations}
% \label{sec:3d_transformations}
The proposed framework (Fig.~\ref{fig:model_3d}) naturally generalizes beyond 2D image reconstruction.  
By extending the MLP decoder to additionally predict a density term \(\sigma\),  
\[
f_\theta(\mathbf{x}) = (\mathbf{c}, \sigma),
\]
the model integrates with a volumetric rendering formulation analogous to \allowbreak NeRF~\cite{mildenhall2021nerf}. In this case, the rendering equation is expressed as
\[
\hat{\mathbf{C}}(\mathbf{r}) = 
\int_{t_n}^{t_f} 
T(t) \, \sigma(\mathbf{r}(t)) \, \mathbf{c}(\mathbf{r}(t)) \, dt,\]
where,
\[T(t) = \exp\!\left(- \int_{t_n}^{t} \sigma(\mathbf{r}(s)) \, ds \right),
\]
\(\mathbf{r}(t) = \mathbf{o} + t \mathbf{d}\) denotes a camera ray with origin \(\mathbf{o}\) and direction \(\mathbf{d}\).

Each Gaussian component is then lifted into three-dimensional space. The local embedding for any query position \(\mathbf{x}\) is computed through the Gaussian-weighted KNN aggregation \(\mathbf{e}_{\mathbf{KNN}}(\mathbf{x}, \G).\) However the KNN utilizes Mahalanobis distance instead of Euclidean distance for the search for nearest neighbors.
As in the 2D formulation, shifting the Gaussian means 
\(\boldsymbol{\mu}_i \mapsto \boldsymbol{\mu}_i'\)
directly modifies the interpolated embedding field 
\(\mathbf{e}_{\mathbf{KNN}}(\mathbf{x}, \G)\),
thereby altering the reconstructed volumetric scene while preserving photometric coherence.

Since the decoder \(f_\theta\) depends solely on the aggregated embeddings rather than absolute coordinates, these transformations yield physically consistent and spatially smooth changes in appearance.  
Consequently, \our{} provides a unified framework for continuous, differentiable editing across both 2D and 3D domains, enabling 2D-to-3D lifting followed by direct 3D scene editing.
%%%%%%%%%%%%%%%%%%%%%%%%%%%%%%%
\section{Experiments}
\label{sec:experiments}
%%%%%%%%%%%%%%%%%%%%%%%%%%%%%%%

\begin{figure}[t]
\centering
    \includegraphics[width=\linewidth]{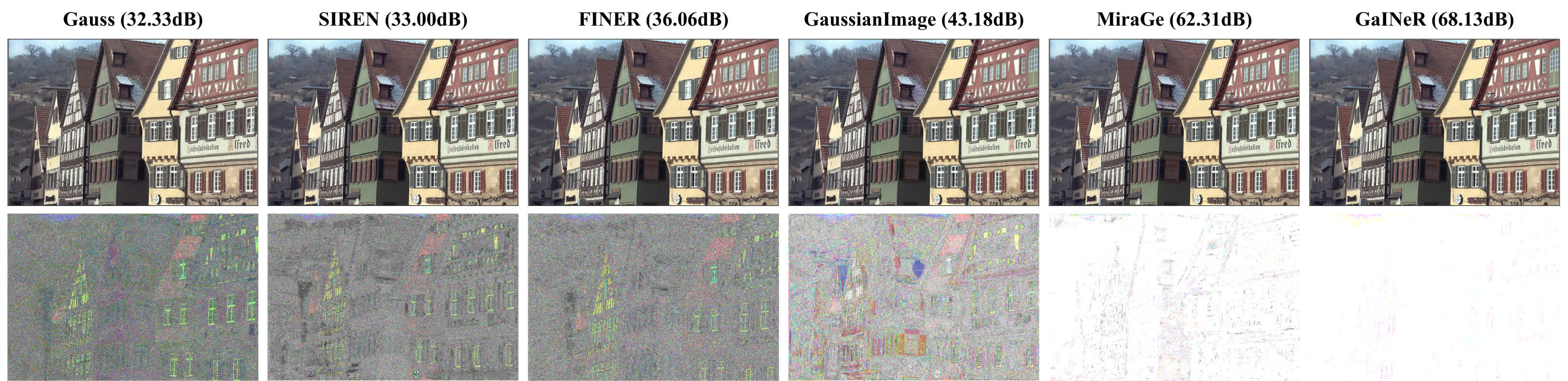}
    \caption{Visual comparison of pixel-wise absolute errors between ground truth and reconstructed images produced by our method and competing approaches. Error magnitudes are contrast-enhanced using gamma correction ($\gamma = 0.2$) for better visibility.}
    \vspace{-0.3cm}
    \label{fig:error_images}
\end{figure}

We evaluated our proposed method across three tasks: image reconstruction, image editing, and 2D-to-3D lifting. The image editing task encompassed both user-driven transformations and physically simulated deformations. For reconstruction, we employed two standard benchmarks: Kodak~\cite{kodak_dataset} and DIV2K ~\cite{Agustsson_2017_CVPR_Workshops}. Additional ablation studies, runtime analysis, and implementation details are provided in the supplementary material. The Kodak dataset contains 24 natural images of resolution \(768 \times 512\), covering diverse scenes such as portraits, landscapes, and textures. 
% The DIV2K validation set involves 2× bicubic downscaling and comprises 100 images with resolutions ranging from 408×1020 to 1020×1020. DIV2K provides 100 high-resolution validation images widely used for assessing image restoration and super-resolution performance.
The DIV2K validation set comprises 100 high-resolution images subjected to 2x bicubic downscaling, with spatial resolutions ranging from 408 x 1020 to 1020 x 1020, and is widely used for evaluating image restoration and super-resolution methods. 

\begin{table}[h]
    \centering
    \small
    \resizebox{0.6\linewidth}{!}{
    \begin{tabular}{lcccc}
        \toprule
        {\textbf{Model}} & \multicolumn{2}{c}{\textbf{Kodak}} & \multicolumn{2}{c}{\textbf{DIV2K}} \\
        \cmidrule(lr){2-3} \cmidrule(lr){4-5}
         & PSNR↑ & MS-SSIM↑ & PSNR↑ & MS-SSIM↑ \\
        \midrule
        WIRE \cite{saragadam2023wire} & 41.47 & 0.9939 & 35.64 & 0.9511 \\
        SIREN \cite{sitzmann2020implicit} & 40.83 & 0.9960 & 36.02 & 0.9568 \\
        I-NGP \cite{muller2022instant} & 43.88 & 0.9976 & 37.06 & 0.9894 \\
        NeuRBF \cite{chen2023neurbfneuralfieldsrepresentation} & 43.78 & 0.9984 & 37.06 & 0.9901 \\
        3DGS \cite{kerbl3Dgaussians} & 44.09 & 0.9991 & 39.12 & 0.9980 \\
        GaussianImage \cite{zhang2024gaussianimage} & 38.93 & 0.9984 & 41.48 & 0.9981 \\
        MiraGe \cite{waczynskamirage} & 59.52 & \textbf{0.9999} & 54.54 & 0.9998 \\
        \textbf{GaINeR (ours)} & \textbf{77.09} & \textbf{0.9999} & \textbf{62.20} & \textbf{0.9999} \\
        \bottomrule
    \end{tabular}
    }
    \caption{Quantitative comparison on the Kodak and DIV2K datasets. Our method achieves the highest reconstruction quality across both datasets in terms of PSNR and MS-SSIM.}
    \label{tab:metrics}
    \vspace{-0.3cm}
\end{table}

\textbf{Quantitative and Qualitative Evaluation}
Table ~\ref{tab:metrics} reports the average PSNR and MS-SSIM values obtained on both datasets. We compared our approach against WIRE \cite{saragadam2023wire}, SIREN \cite{sitzmann2020implicit}, I-NGP \cite{muller2022instant}, NeuRBF \cite{chen2023neurbfneuralfieldsrepresentation}, 3DGS ~\cite{kerbl3Dgaussians}, GaussianImage ~\cite{zhang2024gaussianimage} and MiraGe \cite{waczynskamirage}. In both cases, our method demonstrates a significant improvement in PSNR compared to previous approaches.
Specifically, compared to the strongest existing method (MiraGe), our approach improves the average PSNR from 59.52  to 77.09  on the Kodak dataset, and from 54.54 to 62.20  on DIV2K, while maintaining near-perfect MS-SSIM values. This significant gain indicates that our representation captures image details with much higher fidelity and preserves structural consistency more effectively than prior implicit representations.

% To compare performance, we trained six models (SIREN  \cite{sitzmann2020implicit}, 
% WIRE \cite{saragadam2023wire}, 
% Gauss \cite{ramasinghe2022beyond}, 
% FINER \cite{liu2024finer}, 
% GaussianImage \cite{zhang2024gaussianimage}
% and MiraGe \cite{waczynskamirage}) for 30k iterations each to ensure a fair and consistent comparison.

% As shown in Fig.~\ref{fig:plot_psnr}, our model surpasses all competing methods already from the initial iterations, achieving significantly higher PSNR even at the earliest stages of training. Remarkably, after only about 1–2k iterations (depending on the scene), our method reaches PSNR values higher than those obtained by other approaches after 30k iterations of training.

\begin{figure}[t]
\centering
    \includegraphics[width=\linewidth]{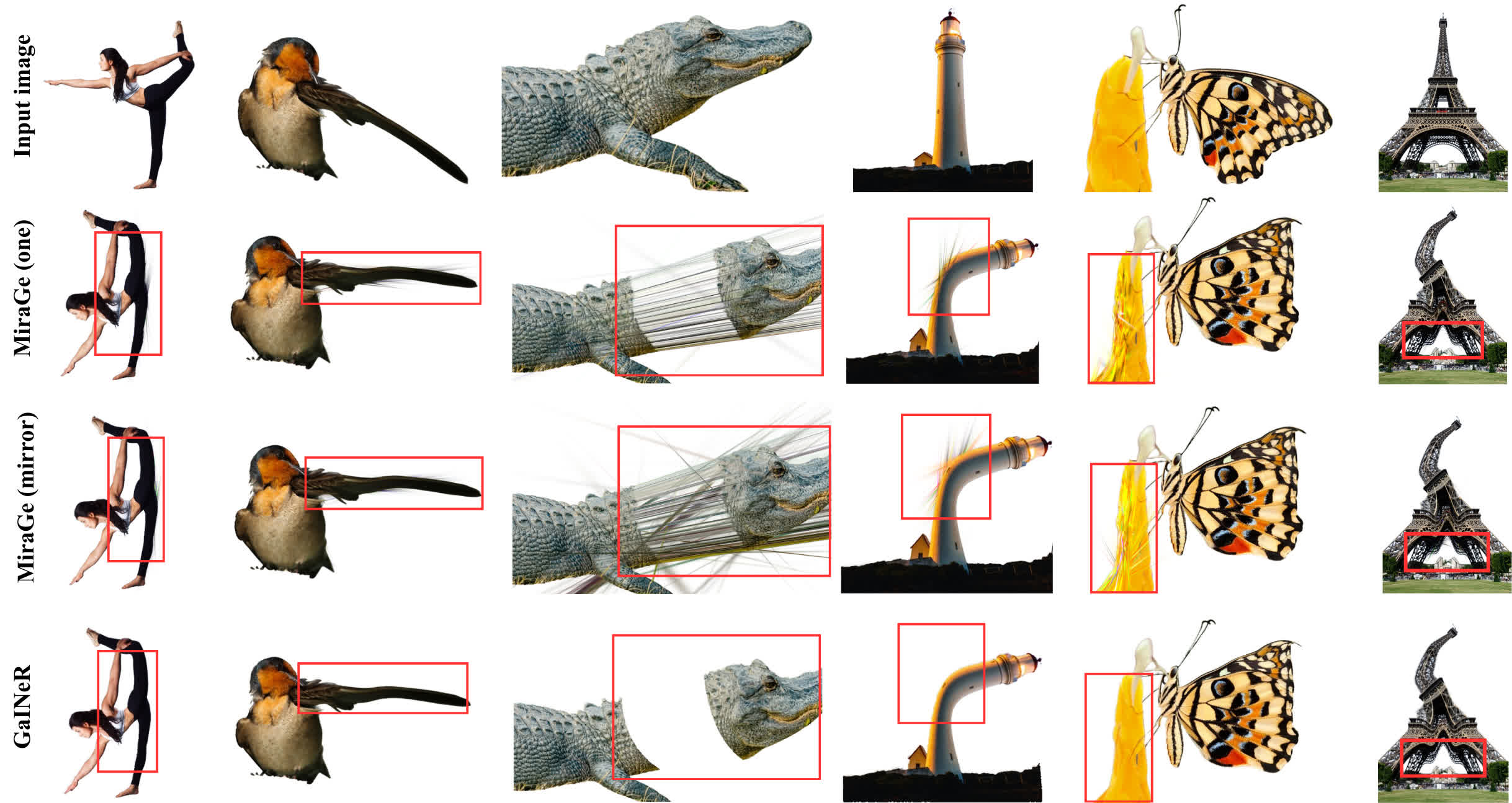}
    \caption{
Comparison of MiraGe and GaINeR under manually applied geometric modifications in Blender. GaINeR maintains stable and realistic results under bending, cutting and stretching, whereas MiraGe exhibits visible artifacts.
    }
    \label{fig:editing_comparisons}
    % \vspace{-0.3cm}
\end{figure}

To ensure consistency across Gaussian-based representations, we evaluate MiraGe, GaussianImage, and our proposed method under an identical training setup. Each model is initialized with 500k Gaussian primitives, which remain fixed throughout optimization (no densification or pruning), and is trained for 30k iterations. As reported in Table \ref{tab:gaussian_comparison} our method achieves the highest accuracy among all evaluated approaches.

\begin{table}[h]
    \centering
    \small
    \resizebox{0.55\linewidth}{!}{
    \begin{tabular}{lcccc}
        \toprule
        {\textbf{Model}} 
        & \multicolumn{2}{c}{\textbf{Kodak}} 
        & \multicolumn{2}{c}{\textbf{DIV2K}} \\
        \cmidrule(lr){2-3} \cmidrule(lr){4-5}
         & PSNR↑ & MS-SSIM↑
         & PSNR↑ & MS-SSIM↑ \\
        \midrule
        GaussianImage \cite{zhang2024gaussianimage}  & 30.46 & 0.9635 & 28.58 & 0.9565 \\
        MiraGe \cite{waczynskamirage}                & 62.49 & \textbf{0.9999} & 58.45 & \textbf{0.9999} \\
        \textbf{GaINeR (ours)}                       & \textbf{77.09} & \textbf{0.9999} & \textbf{62.20} & \textbf{0.9999} \\
        \bottomrule
    \end{tabular}
    }
    \caption{Comparison with Gaussian-based approaches on the Kodak and DIV2K datasets. Our method achieves the highest reconstruction quality in terms of PSNR and MS-SSIM.}
    \label{tab:gaussian_comparison}
    \vspace{-0.3cm}
\end{table}

% \subsection{Qualitative comparisons}

Since the PSNR values across all methods are exceptionally high, the visual differences are barely perceptible to the human eye.  
Therefore, we compute the \textit{absolute error} for each pixel and visualize the results as error maps, shown in Fig.~\ref{fig:error_images}.  
These error visualizations reveal that \our{} method produces almost no noticeable errors compared to other approaches, demonstrating its superior reconstruction accuracy and stability.

\textbf{Editing comparisons}
We conducted a series of experiments on images from the DIV2K dataset, exploring various types of spatial modifications. For consistency across methods, we evaluate MiraGe using two camera configurations: one and mirror. The one camera setup corresponds to the same viewpoint configuration employed in our \our{} model, ensuring direct comparability. The mirror camera provides a reflective viewpoint that captures mirrored scene geometry and appearance.
As illustrated in Fig.~\ref{fig:editing_comparisons}, we evaluate the visual effects of bending, stretching, cutting and displacing Gaussian components.  
MiraGe exhibits severe distortions under these transformations, primarily due to its reliance on Gaussian primitives for direct rendering.  
In contrast, our \our{} framework produces visually consistent and realistic results across all conditions, as it utilizes Gaussian components solely for encoding rather than for rendering.  
% We additionally evaluate the behavior of the model under simple scale manipulations by enlarging and shrinking the main object in the scene. As shown in Fig.~\ref{fig:up_and_down}, both upscaling and downscaling preserve the structural continuity of the object and do not introduce visible artifacts, resulting in visually consistent edits.

\begin{figure}[t]
\centering
\includegraphics[width=\linewidth]{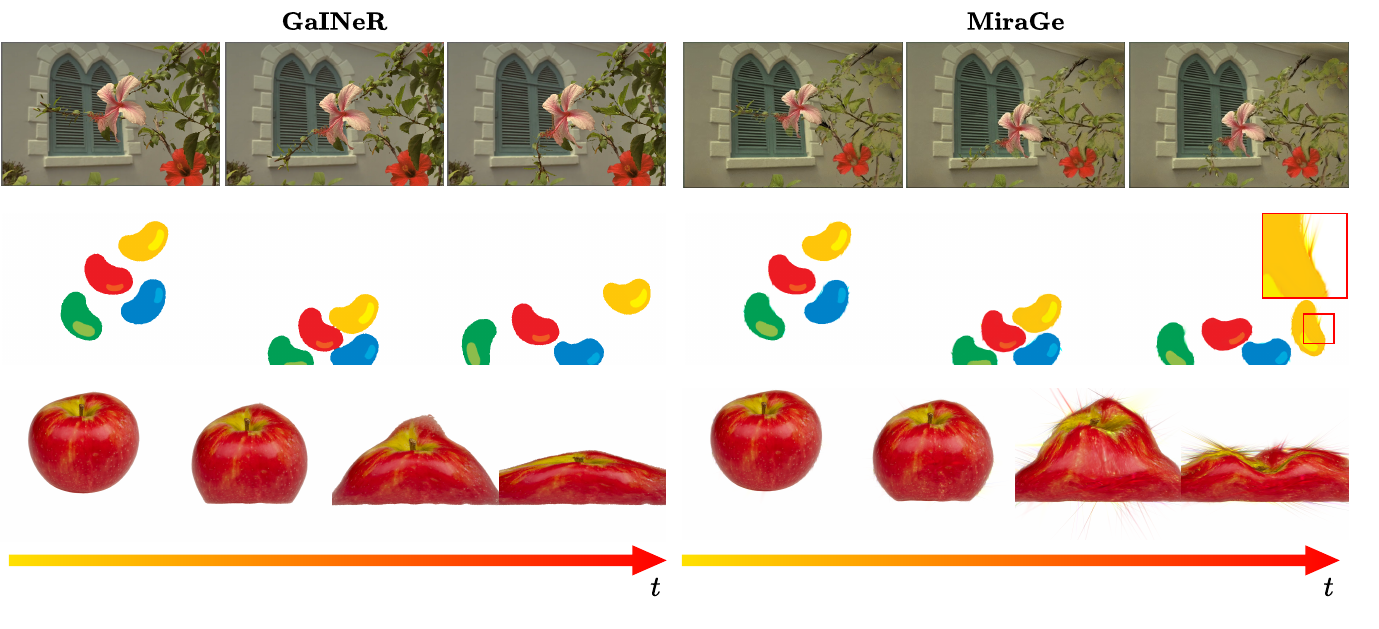}
\caption{
Qualitative comparison of physics-based animations.
\textbf{Elastic Flower:} MiraGe's render exhibits transparencies, while ours remains opaque.
\textbf{Elastic Beans:} MiraGe produces spiky boundary artifacts, which are absent in our result.
\textbf{Sand Apple:} As the object collapses, MiraGe's rendering shows structural artifacts, unlike \our{}.
}
\vspace{-0.3cm}
\label{fig:physics_comparison}
\end{figure}

% \begin{figure}[t]
% \centering
% \includegraphics[width=\linewidth]{ECCV/figures/downscaled_upscaled.png}
% \caption{
% GaINeR tested under scale-based geometric modifications, including object enlargement and reduction. In all cases, the
% method yields stable and visually coherent results.
% }
% \vspace{-0.3cm}
% \label{fig:up_and_down}
% \end{figure}

\begin{figure}[t]
\centering
\includegraphics[width=\linewidth]{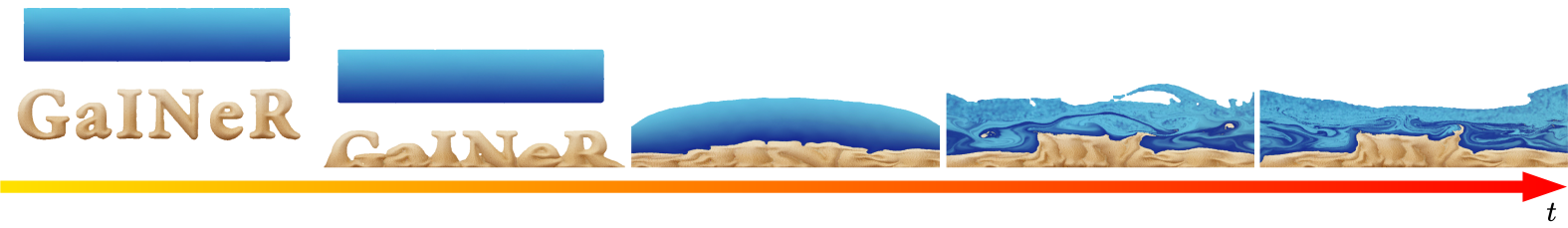}
\caption{
    Physical simulation of complex multi-material interactions using \our{}. By treating learned Gaussian centers as material points in a Material Point Method engine, we simulate granular dynamics (sand) and fluid mechanics (water) simultaneously. Our continuous MLP decoder maintains cohesive surfaces, fluid-like blending, and high-frequency textures even under severe topological deformations over time.
}
\vspace{-0.3cm}
\label{fig:sim_gainer}
\end{figure}

\textbf{Physics comparisons}
To demonstrate applicability in dynamic scenarios, we treat the centers of our learned Gaussians as material points in a Material Point Method (MPM) simulation driven by the Taichi Elements~\cite{taichi_elements} engine. Fig.~\ref{fig:physics_comparison} compares our method against MiraGe, with full animations provided in the supplementary material. While MiraGe's discrete representation produces geometric artifacts and spiky boundaries under stress, \our{} renders coherent surfaces even under significant deformation. Since our MLP decoder is conditioned on the geometric structure, it can robustly interpolate visual information from sparse or distorted particle distributions, avoiding the visual discontinuities inherent to discrete representations. We further showcase this robustness in Fig.~\ref{fig:sim_gainer}, where \our{} successfully handles complex, multi-material interactions, such as fluid mechanics and granular dynamics, maintaining continuous, high-fidelity textures despite extreme topological changes over time.

\textbf{2D-to-3D Lifting} We performed a series of experiments demonstrating the capability of \our{} to lift 2D representations into 3D space.  
As illustrated in Fig.~\ref{fig:2d_to_3d}, \our{} reconstructs realistic and geometrically consistent 3D models by leveraging the spatially coherent nature of Gaussian embeddings and the KNN-based aggregation mechanism extended to three dimensions.  

We evaluated our method on various objects from the LLFF~\cite{mildenhall2019llff} and DTU~\cite{aanaes2012dtu} datasets (see also Fig.~\ref{fig:model_3d}).  
Due to the continuity of the Gaussian representation in 3D, \our{} also produces accurate depth maps for rendered views.  
For these experiments, depth supervision was provided using depth maps estimated by Depth-Pro~\cite{bochkovskiy2025depthpro}, which were normalized to ensure consistent metric ranges across datasets.  

Overall, \our{} demonstrates strong performance in reconstructing and rendering 3D geometry from 2D inputs, achieving realistic depth perception and spatial coherence.

% Image upscaling aims to reconstruct high-resolution images from low-resolution inputs while preserving fine details and structural consistency. This task becomes particularly challenging at high magnification factors, where artifacts and over-smoothing effects often appear. Figure \ref{fig:upscaling} presents a qualitative comparison of our method, GaINeR, with 3D Gaussian Splatting and MiraGe at ×8, ×10, and ×12 scaling factors.
% GaINeR is based on an implicit neural representation (INR), which allows stable generation of high-resolution outputs without relying on fixed discrete grids. As shown in the figure, GaINeR preserves fine textures and sharp edges even at extreme magnifications. In contrast, 3D Gaussian Splatting and MiraGe exhibit visible degradation artifacts as the scale increases, in particular Gaussian-like artifacts and over-smoothed regions. All experiments were conducted on the DIV2K dataset. During training, the input images were downscaled by a factor of eight with respect to their original resolution.
\textbf{Upscaling}
\begin{figure}[t]
\centering
    \includegraphics[width=\linewidth]{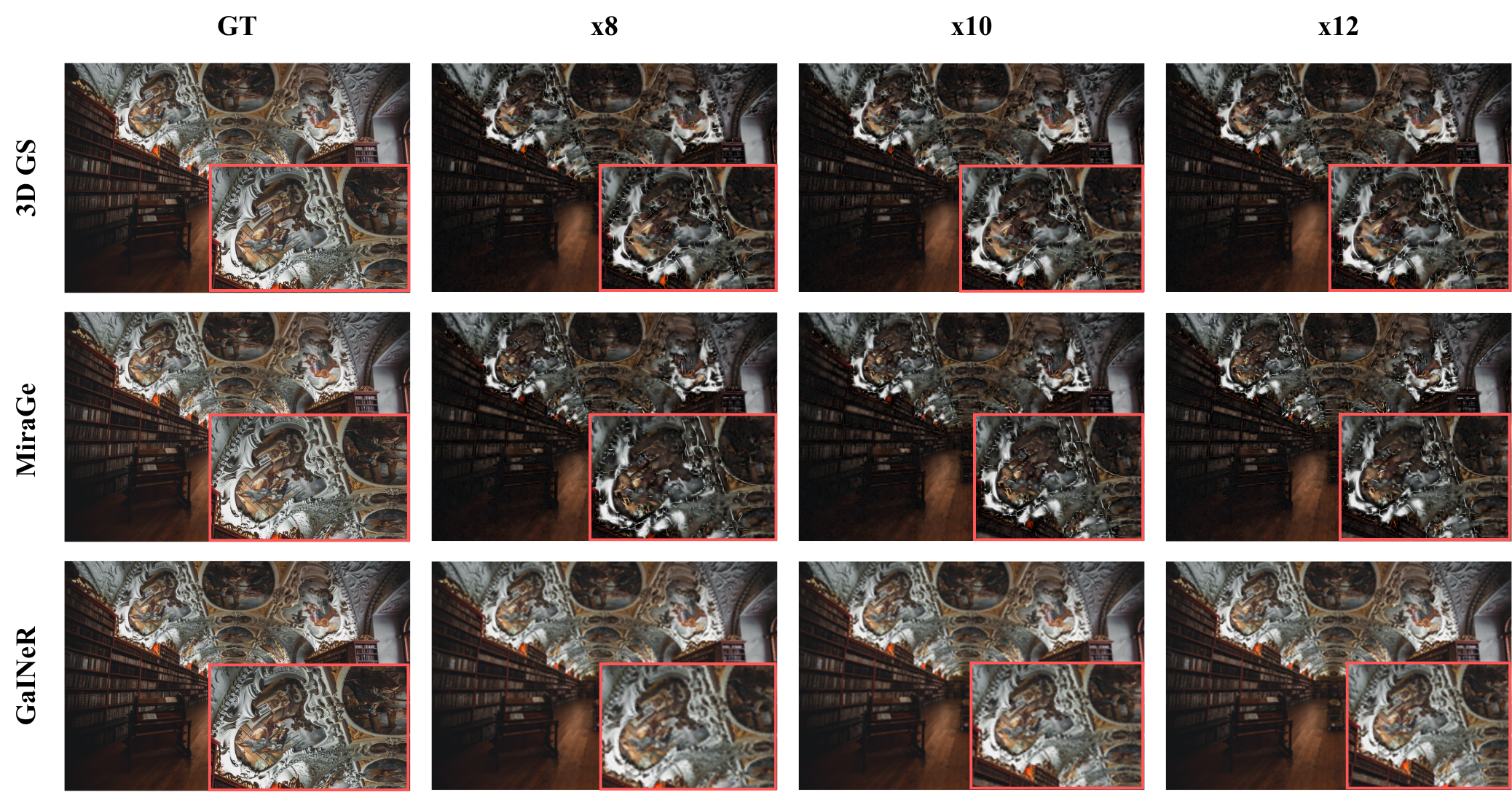}
    \caption{
    Visual comparison of GaINeR with 3D Gaussian Splatting and MiraGe at ×8, ×10, and ×12 scaling factors. Our method maintains higher perceptual quality, whereas competing approaches introduce Gaussian artifacts.
    }
    \label{fig:upscaling}
    % \vspace{-0.3cm}
\end{figure}
Image upscaling aims to reconstruct high-resolution images from low-resolution inputs while preserving fine details and structural consistency, a task that becomes particularly challenging at high magnification factors where artifacts and over-smoothing often appear. Fig.~\ref{fig:upscaling} shows a qualitative comparison of GaINeR with 3D Gaussian Splatting and MiraGe at ×8, ×10, and ×12 scales.
As an implicit neural representation (INR), GaINeR generates high-resolution outputs without relying on fixed discrete grids. As illustrated in the figure, it preserves fine textures and sharp edges even at extreme magnifications, while 3D Gaussian Splatting and MiraGe exhibit visible degradation artifacts, including Gaussian-like patterns and over-smoothed regions. All experiments were conducted on the DIV2K dataset, where the input images were obtained by bicubic downscaling of the original images by a factor of 8.

%%%%%%%%%%%%%%%%%%%%%%%%%%%%%%%
\section{Conclusion}
%%%%%%%%%%%%%%%%%%%%%%%%%%%%%%%

In this work, we introduced \our{}, a geometry-aware implicit representation that combines the continuous flexibility of neural networks with the structured expressiveness of learnable Gaussian embeddings.  
By encoding spatial coordinates through Gaussian-weighted KNN aggregation, \our{} achieves smooth, spatially coherent reconstructions while enabling 2D image manipulations.

Our framework bridges coordinate-based MLPs with explicit geometric priors, yielding a unified model capable of high-fidelity reconstruction and photometrically consistent transformations.  
Comprehensive experiments demonstrate that \our{} surpasses prior implicit representations such as SIREN, WIRE, and MiraGe in both quantitative accuracy and visual quality, while offering greater interpretability through its Gaussian-based encoding. In addition, the continuous formulation of \our{} naturally supports arbitrary-resolution image synthesis, enabling seamless super-resolution and upscaling without retraining.

Furthermore, we extend the formulation to 2D-to-3D lifting, where the learned Gaussian embeddings are projected into three-dimensional space to reconstruct depth-aware geometry and novel viewpoints from a single image.  
This generalization highlights the potential of Gaussian-encoded neural fields for bridging 2D perception and 3D reasoning, paving the way for future research in continuous, geometry-aware scene representations.

\textbf{Limitations} While \our{} enables 3D lifting, its geometry is bounded by the quality of the off-the-shelf monocular depth estimator. Errors in depth prediction can lead to geometric artifacts.

\bibliographystyle{splncs04}
\bibliography{main}

@String(CVPR  = {IEEE Conf. Comput. Vis. Pattern Recog.})

@String(ICCV  = {Int. Conf. Comput. Vis.})

@String(ICLR  = {Int. Conf. Learn. Represent.})

@String(TOG   = {ACM Trans. Graph.})

@String(CVPR  = {CVPR})

@String(ICCV  = {ICCV})

@String(ICLR  = {ICLR})

@String(TOG   = {ACM TOG})

@incollection{takikawa2023neural,
  title={Neural Fields for Visual Computing: SIGGRAPH 2023 Course},
  author={Takikawa, Towaki and Saito, Shunsuke and Tompkin, James and Sitzmann, Vincent and Sridhar, Srinath and Litany, Or and Yu, Alex},
  booktitle={ACM SIGGRAPH 2023 Courses},
  pages={1--227},
  year={2023}
}

@article{sitzmann2020implicit,
  title={Implicit neural representations with periodic activation functions},
  author={Sitzmann, Vincent and Martel, Julien and Bergman, Alexander and Lindell, David and Wetzstein, Gordon},
  journal={Advances in neural information processing systems},
  volume={33},
  pages={7462--7473},
  year={2020}
}

@inproceedings{klocek2019hypernetwork,
  title={Hypernetwork functional image representation},
  author={Klocek, Sylwester and Maziarka, {\L}ukasz and Wo{\l}czyk, Maciej and Tabor, Jacek and Nowak, Jakub and {\'S}mieja, Marek},
  booktitle={International Conference on Artificial Neural Networks},
  pages={496--510},
  year={2019},
  organization={Springer}
}

@inproceedings{ramasinghe2022beyond,
  title={Beyond periodicity: Towards a unifying framework for activations in coordinate-mlps},
  author={Ramasinghe, Sameera and Lucey, Simon},
  booktitle={European Conference on Computer Vision},
  pages={142--158},
  year={2022},
  organization={Springer}
}

@inproceedings{saragadam2023wire,
  title={Wire: Wavelet implicit neural representations},
  author={Saragadam, Vishwanath and LeJeune, Daniel and Tan, Jasper and Balakrishnan, Guha and Veeraraghavan, Ashok and Baraniuk, Richard G},
  booktitle={Proceedings of the IEEE/CVF Conference on Computer Vision and Pattern Recognition},
  pages={18507--18516},
  year={2023}
}

@inproceedings{liu2024finer,
  title={Finer: Flexible spectral-bias tuning in implicit neural representation by variable-periodic activation functions},
  author={Liu, Zhen and Zhu, Hao and Zhang, Qi and Fu, Jingde and Deng, Weibing and Ma, Zhan and Guo, Yanwen and Cao, Xun},
  booktitle={Proceedings of the IEEE/CVF Conference on Computer Vision and Pattern Recognition},
  pages={2713--2722},
  year={2024}
}

@article{tancik2020fourier,
  title={Fourier features let networks learn high frequency functions in low dimensional domains},
  author={Tancik, Matthew and Srinivasan, Pratul and Mildenhall, Ben and Fridovich-Keil, Sara and Raghavan, Nithin and Singhal, Utkarsh and Ramamoorthi, Ravi and Barron, Jonathan and Ng, Ren},
  journal={Advances in neural information processing systems},
  volume={33},
  pages={7537--7547},
  year={2020}
}

@inproceedings{waczynskamirage,
  title={MiraGe: Editable 2D Images using Gaussian Splatting},
  author={Waczynska, Joanna and Szczepanik, Tomasz and Borycki, Piotr and Tadeja, Slawomir and Bohn{\'e}, Thomas and Spurek, Przemys{\l}aw},
  booktitle={Forty-second International Conference on Machine Learning},
  year={2025}
}

@inproceedings{kaniafresh,
  title={FreSh: Frequency Shifting for Accelerated Neural Representation Learning},
  author={Kania, Adam and Mihajlovic, Marko and Prokudin, Sergey and Tabor, Jacek and Spurek, Przemys{\l}aw},
  booktitle={The Thirteenth International Conference on Learning Representations},
  year={2025}
}

@inproceedings{tewari2022advances,
  title={Advances in neural rendering},
  author={Tewari, Ayush and Thies, Justus and Mildenhall, Ben and Srinivasan, Pratul and Tretschk, Edgar and Yifan, Wang and Lassner, Christoph and Sitzmann, Vincent and Martin-Brualla, Ricardo and Lombardi, Stephen and others},
  booktitle={Computer Graphics Forum},
  volume={41},
  number={2},
  pages={703--735},
  year={2022},
  organization={Wiley Online Library}
}

@inproceedings{chen2022videoinr,
  title={Videoinr: Learning video implicit neural representation for continuous space-time super-resolution},
  author={Chen, Zeyuan and Chen, Yinbo and Liu, Jingwen and Xu, Xingqian and Goel, Vidit and Wang, Zhangyang and Shi, Humphrey and Wang, Xiaolong},
  booktitle={Proceedings of the IEEE/CVF Conference on Computer Vision and Pattern Recognition},
  pages={2047--2057},
  year={2022}
}

@inproceedings{park2019deepsdf,
  title={Deepsdf: Learning continuous signed distance functions for shape representation},
  author={Park, Jeong Joon and Florence, Peter and Straub, Julian and Newcombe, Richard and Lovegrove, Steven},
  booktitle={Proceedings of the IEEE/CVF conference on computer vision and pattern recognition},
  pages={165--174},
  year={2019}
}

@article{mildenhall2021nerf,
  title={Nerf: Representing scenes as neural radiance fields for view synthesis},
  author={Mildenhall, Ben and Srinivasan, Pratul P and Tancik, Matthew and Barron, Jonathan T and Ramamoorthi, Ravi and Ng, Ren},
  journal={Communications of the ACM},
  volume={65},
  number={1},
  pages={99--106},
  year={2021},
  publisher={ACM New York, NY, USA}
}

@inproceedings{lin2021barf,
  title={Barf: Bundle-adjusting neural radiance fields},
  author={Lin, Chen-Hsuan and Ma, Wei-Chiu and Torralba, Antonio and Lucey, Simon},
  booktitle={Proceedings of the IEEE/CVF international conference on computer vision},
  pages={5741--5751},
  year={2021}
}

@inproceedings{chen2022tensorf,
  title={Tensorf: Tensorial radiance fields},
  author={Chen, Anpei and Xu, Zexiang and Geiger, Andreas and Yu, Jingyi and Su, Hao},
  booktitle={European conference on computer vision},
  pages={333--350},
  year={2022},
  organization={Springer}
}

@article{muller2022instant,
  title={Instant neural graphics primitives with a multiresolution hash encoding},
  author={M{\"u}ller, Thomas and Evans, Alex and Schied, Christoph and Keller, Alexander},
  journal={ACM transactions on graphics (TOG)},
  volume={41},
  number={4},
  pages={1--15},
  year={2022},
  publisher={ACM New York, NY, USA}
}

@article{sitzmann2020metasdf,
  title={Metasdf: Meta-learning signed distance functions},
  author={Sitzmann, Vincent and Chan, Eric and Tucker, Richard and Snavely, Noah and Wetzstein, Gordon},
  journal={Advances in Neural Information Processing Systems},
  volume={33},
  pages={10136--10147},
  year={2020}
}

@inproceedings{yang2023freenerf,
  title={Freenerf: Improving few-shot neural rendering with free frequency regularization},
  author={Yang, Jiawei and Pavone, Marco and Wang, Yue},
  booktitle={Proceedings of the IEEE/CVF conference on computer vision and pattern recognition},
  pages={8254--8263},
  year={2023}
}

@inproceedings{barron2021mip,
  title={Mip-nerf: A multiscale representation for anti-aliasing neural radiance fields},
  author={Barron, Jonathan T and Mildenhall, Ben and Tancik, Matthew and Hedman, Peter and Martin-Brualla, Ricardo and Srinivasan, Pratul P},
  booktitle={Proceedings of the IEEE/CVF international conference on computer vision},
  pages={5855--5864},
  year={2021}
}

@inproceedings{zhang2024gaussianimage,
  title={Gaussianimage: 1000 fps image representation and compression by 2d gaussian splatting},
  author={Zhang, Xinjie and Ge, Xingtong and Xu, Tongda and He, Dailan and Wang, Yan and Qin, Hongwei and Lu, Guo and Geng, Jing and Zhang, Jun},
  booktitle={European Conference on Computer Vision},
  pages={327--345},
  year={2024},
  organization={Springer}
}

@Article{kerbl3Dgaussians,
      author       = {Kerbl, Bernhard and Kopanas, Georgeios and Leimk{\"u}hler, Thomas and Drettakis, George},
      title        = {3D Gaussian Splatting for Real-Time Radiance Field Rendering},
      journal      = {ACM Transactions on Graphics},
      number       = {4},
      volume       = {42},
      month        = {July},
      year         = {2023},
      url          = {https://repo-sam.inria.fr/fungraph/3d-gaussian-splatting/}
}

@misc{chen2023neurbfneuralfieldsrepresentation,
      title={NeuRBF: A Neural Fields Representation with Adaptive Radial Basis Functions}, 
      author={Zhang Chen and Zhong Li and Liangchen Song and Lele Chen and Jingyi Yu and Junsong Yuan and Yi Xu},
      year={2023},
      eprint={2309.15426},
      archivePrefix={arXiv},
      primaryClass={cs.CV},
      url={https://arxiv.org/abs/2309.15426}, 
}

@misc{taichi_elements,
  author = {{Taichi Graphics}},
  title  = {Taichi Elements: A High-Performance Multi-Material Continuum Physics Engine},
  year   = {2020},
  note   = {\url{https://github.com/taichi-dev/taichi_elements}},
  howpublished = {GitHub repository}
}

@misc{kodak_dataset,
  author = {Eastman Kodak Company},
  title  = {Kodak Lossless True Color Image Suite},
  year   = {1991},
  note   = {Available at \url{https://r0k.us/graphics/kodak/}},
  howpublished = {Web page}
}

@inproceedings{kirillov2023segment,
  author = {Kirillov, Alexander and Mintun, Eric and Ravi, Nikhila and Mao, Hanzi and Rolland, Chloé and Gustafson, Laura and Xiao, Tete and Whitehead, Spencer and Berg, Alexander C. and Lo, Wan-Yen and Dollár, Piotr and Girshick, Ross B.},
  booktitle = {ICCV},
  isbn = {979-8-3503-0718-4},
  pages = {3992-4003},
  publisher = {IEEE},
  title = {Segment Anything.},
  year = 2023
}

@inproceedings{bochkovskiy2025depthpro,
  author = {Bochkovskiy, Alexey and Delaunoy, Amaël and Germain, Hugo and Santos, Marcel and Zhou, Yichao and Richter, Stephan R. and Koltun, Vladlen},
  booktitle = {ICLR},
  publisher = {OpenReview.net},
  title = {Depth Pro: Sharp Monocular Metric Depth in Less Than a Second.},
  year = 2025
}

@InProceedings{Agustsson_2017_CVPR_Workshops,
	author = {Agustsson, Eirikur and Timofte, Radu},
	title = {NTIRE 2017 Challenge on Single Image Super-Resolution: Dataset and Study},
	booktitle = {The IEEE Conference on Computer Vision and Pattern Recognition (CVPR) Workshops},
	month = {July},
	year = {2017}
}

@article{mildenhall2019llff,
  title={Local Light Field Fusion: Practical View Synthesis with Prescriptive Sampling Guidelines},
  author={Ben Mildenhall and Pratul P. Srinivasan and Rodrigo Ortiz-Cayon and Nima Khademi Kalantari and Ravi Ramamoorthi and Ren Ng and Abhishek Kar},
  journal={ACM Transactions on Graphics (TOG)},
  year={2019},
}

@article{aanaes2012dtu,
  title={Interesting Interest Points},
  author={Aan{\ae}s, H. and Dahl, A.L. and Steenstrup Pedersen, K.},
  journal={International Journal of Computer Vision},
  pages={18--35},
  year={2012},
  volume={97},
  publisher={Springer}
}

\clearpage
\appendix

% \documentclass[runningheads]{llncs}

% % \usepackage[review,year=2026,ID=2019]{eccv}

% \usepackage{graphicx}
% \usepackage{booktabs}
% \usepackage{eccvabbrv}
% \usepackage{hyperref}

\newcommand{\cred}[1]{{\color{red}#1}}

\def\our{GaINeR}
\def\fullname{Geometry-Aware Implicit Network Representation}

% \begin{document}

\title{GaINeR: Geometry-Aware Implicit Neural Representation for Image Editing\\
Supplementary Material}

\author{
Weronika Jakubowska\inst{1}$^*$\and
Mikołaj Zieliński\inst{2}$^*$\and
Rafał Tobiasz\inst{3,4}$^*$\and
Krzysztof Byrski\inst{4}\and
Maciej Zięba\inst{1}\and
Dominik Belter\inst{2}\and
Przemysław Spurek\inst{3,4}
}
% \authorrunning{Jakubowska et al.}

\institute{
Wrocław University of Science and Technology \and
Poznań University of Technology \and
IDEAS Research Institute \and
Jagiellonian University\\[6pt]
$^*$Equal contribution\\
\email{weronika.jakubowska@pwr.edu.pl}
}

\maketitle

\appendix

\section{Extended Experimental Analysis}
This section presents additional experiments and analyses that complement the main results. Specifically, we provide detailed error comparisons, further editing examples, physics-inspired evaluations, novel 2D-to-3D transformations, computational efficiency comparisons (size and time) and an extensive ablation study. Together, these extended results provide a broader view of the method’s capabilities and behavior.

\subsection{Error Comparisons}

\begin{figure}[]
\centering
    \includegraphics[width=\linewidth]{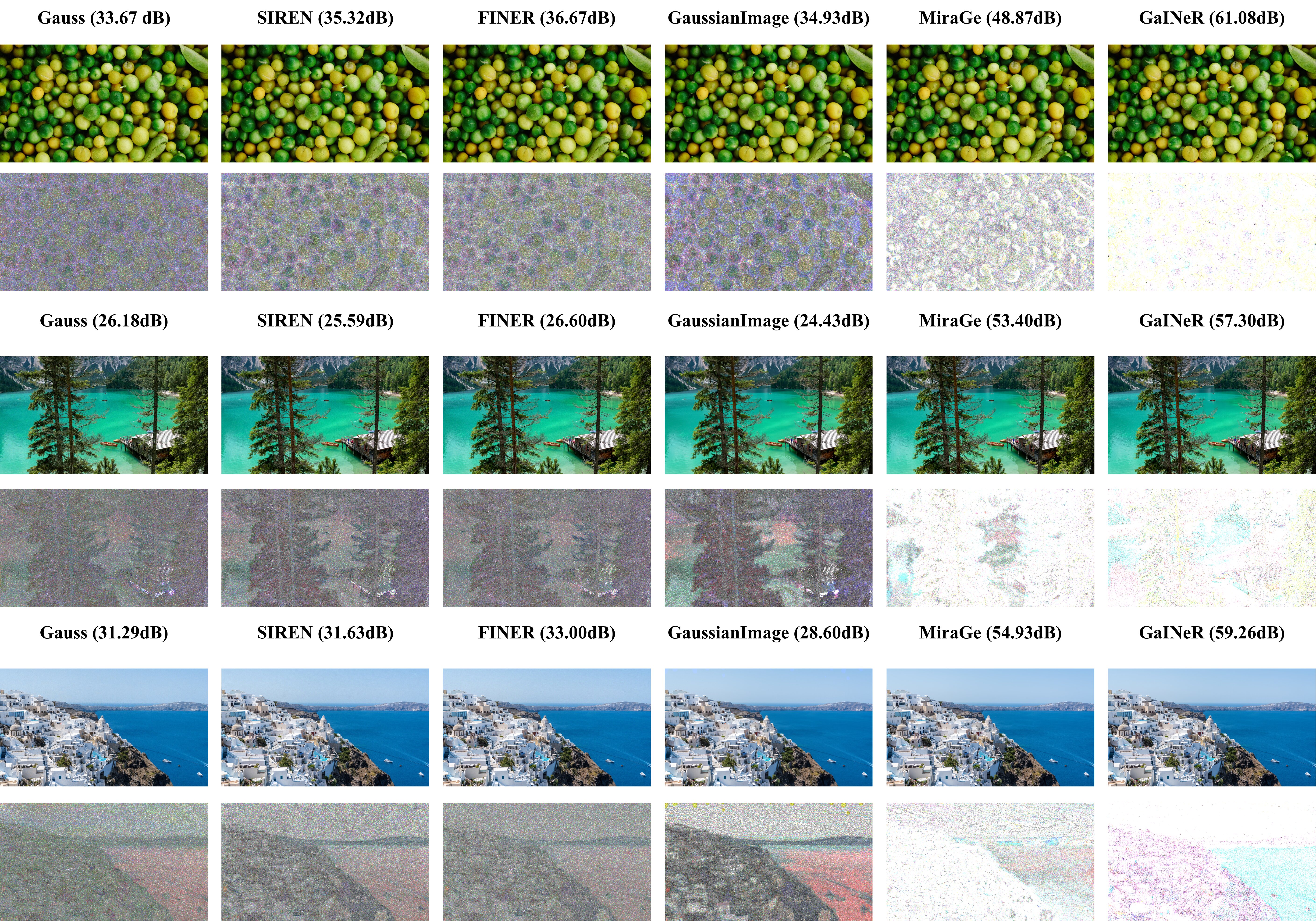}
    \caption{Visual comparison of pixel-wise absolute errors between ground truth and reconstructed images produced by our method and competing approaches. Error magnitudes are contrast-enhanced using gamma correction ($\gamma = 0.2$) for better visibility.}
    
    \label{fig:error_images_div2k}
\end{figure}

In this section, we provide additional examples illustrating the absolute pixel-wise error produced by our method compared to baseline approaches (see Fig. \ref{fig:error_images_div2k}).  
The visualizations highlight areas where reconstruction discrepancies are most pronounced and allow for a detailed qualitative assessment of image fidelity.  
We present results on several images from the DIV2K validation set \cite{Agustsson_2017_CVPR_Workshops}, which we used for evaluation.  
These examples demonstrate that our model consistently produces minimal reconstruction errors, preserving fine details and textures even in challenging regions.

\subsection{Editing Comparisons}
\begin{figure}[]
\centering
\includegraphics[width=\linewidth]{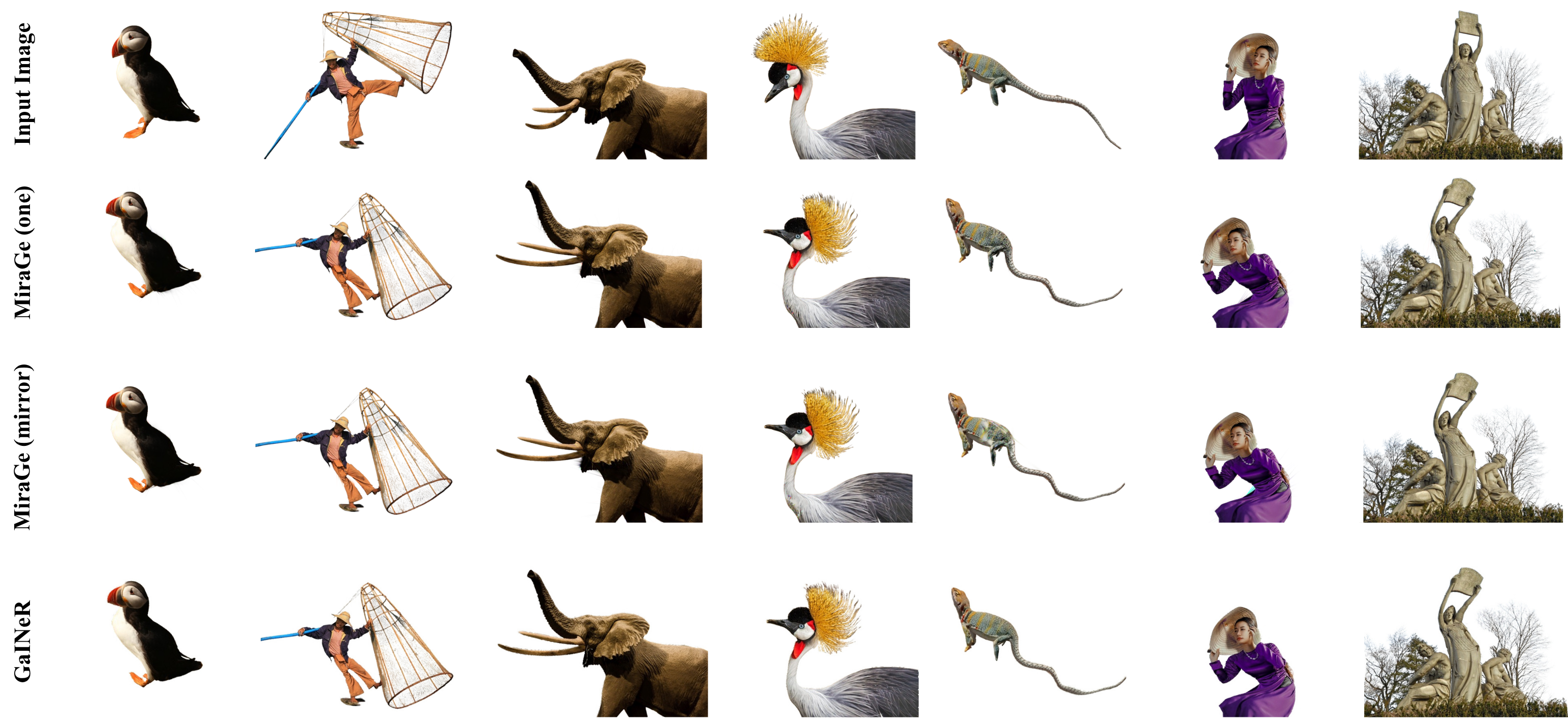}
\caption{
Extended qualitative assessment of MiraGe \cite{waczynskamirage} and GaINeR under manually introduced geometric transformations. MiraGe shows artifacts in several cases: around the woman’s abdomen, near the elephant’s neck when using the mirror camera, and along the body of the lizard and the neck of the gray crowned crane under the same camera setting.
}
\label{fig:more_edits}
\end{figure}

In Fig.~\ref{fig:more_edits}, we provide additional qualitative comparisons illustrating how the models respond to a broader range of manual geometric modifications. Across most scenarios, \our{} maintains stable geometry and appearance, producing visually coherent results even under stronger or more irregular deformations. MiraGe \cite{waczynskamirage} exhibits structural inconsistencies under such modifications.

\begin{figure*}[]
\centering
\includegraphics[width=\linewidth]{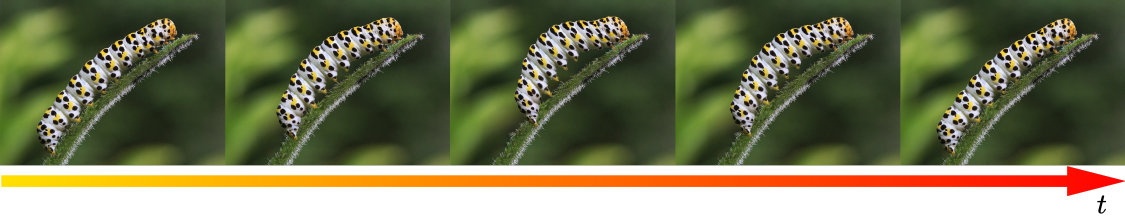}
\caption{
Sequences generated in Blender through frame-wise geometric transformations demonstrate that the method maintains stable structure and visual consistency throughout the animation.
}
\label{fig:animation}
\end{figure*}

Beyond static manual edits, the editing pipeline also supports generating animations, enabling controlled simulations of various object motions. By applying frame-wise geometric transformations, we can assess the temporal stability of the reconstruction process. As illustrated in Fig.~\ref{fig:animation}, \our{} maintains structural coherence and produces artifact-free frames throughout the animated sequence, demonstrating its robustness under continuous, motion-induced transformations.

\subsection{Physics Evaluations}

We further investigate the dynamic capabilities of \our{} through two distinct simulation scenarios powered by the Taichi Elements engine~\cite{taichi_elements}. These experiments demonstrate the versatility of our explicit Gaussian embeddings when utilized as physical particles.

\begin{figure*}[]
\centering
\includegraphics[width=\linewidth]{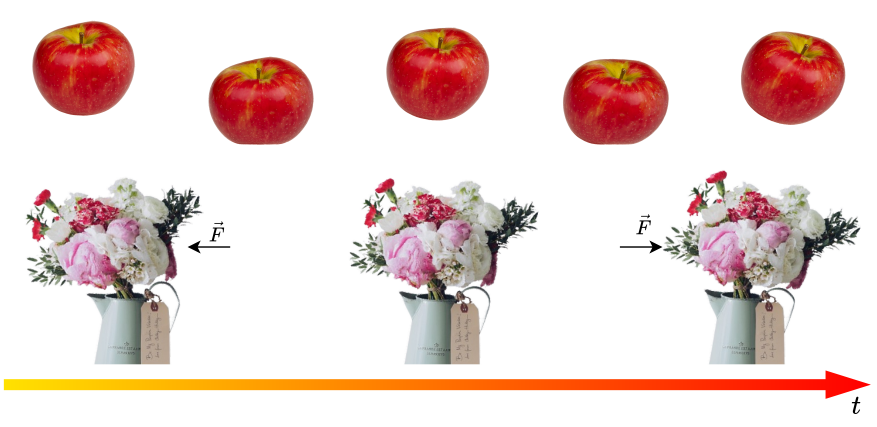}
\caption{
Visualizations of physics-based simulations.
\textbf{Top:} An elastic apple deforming under constant gravity upon impact.
\textbf{Middle:} A bouquet subjected to oscillating lateral forces, demonstrating structural flexibility.
}
\label{fig:simulations_supplement}
\end{figure*}

First, we validate baseline elastic deformation under constant gravity (Fig.~\ref{fig:simulations_supplement}, top), ensuring that the learned representation maintains volumetric consistency during collision. Second, we introduce temporally varying external forces to a complex structure (Fig.~\ref{fig:simulations_supplement}, middle), demonstrating that the decoder produces artifact-free renderings even under oscillating bending deformations. 

\subsection{2D-to-3D Transformations}

In the main manuscript, we established the capability of our framework to lift implicit 2D image representations into navigable 3D environments. Here, we provide additional qualitative results to further validate the robustness of this approach.

Fig.~\ref{fig:2d_to_3d_objects} showcases the 2D-to-3D lifting process applied to complex, isolated objects. The novel view renders demonstrate that \our{} successfully extrudes flat 2D topology into coherent 3D volumes. Importantly, it preserves intricate geometric details such as overlapping petals and complex contours while maintaining strict photometric consistency without introducing severe stretching artifacts.

Furthermore, Fig.~\ref{fig:2d_to_3d_paintings} illustrates the versatility of our framework on full-frame scenes, including classical artworks and dense landscapes. As the camera trajectory deviates from the original frontal viewpoint, the synthesized renders exhibit a highly realistic parallax effect. These results confirm that our model maintains textural coherence and correct spatial relationships across varying depth planes, all derived entirely from a single unposed image.

\begin{figure}[]
\centering
    \includegraphics[width=\linewidth]{figures_suppl/2d_to_3d_objects.pdf}
    \caption{Qualitative results of 2D-to-3D lifting on isolated objects. By leveraging monocular depth maps to spatially translate the learned Gaussian embeddings along the Z-axis, GaINeR accurately extrudes flat 2D representations into coherent 3D volumes.}
    
    \label{fig:2d_to_3d_objects}
\end{figure}

\begin{figure}[]
\centering
    \includegraphics[width=\linewidth]{figures_suppl/2d_to_3d_paintings.pdf}
    \caption{Immersive 2D-to-3D lifting on full-frame paintings and landscapes. Driven by the corresponding continuous depth gradients, GaINeR transforms traditional flat artworks into navigable volumetric scenes.}
    
    \label{fig:2d_to_3d_paintings}
\end{figure}

\subsection{Ablation study}
We conduct an ablation study on the KODAK dataset to evaluate the effect of key design parameters in \our{}.  
Specifically, we analyze the impact of the number of Gaussian components ($N_G$), neighborhood size ($N_{KNN}$), search radius ($r$), hashgrid resolution ($\mathcal{H}_{res}$), and hash map capacity ($H_{size}$) on reconstruction quality.  
Quantitative results are summarized in Table~\ref{tab:ablation}.
        \begin{table}[]
        \centering
        \footnotesize
        \begin{tabular}{ccccc|cc}
        \toprule
        \multicolumn{5}{c|}{\textbf{Parameter values}} & \multicolumn{2}{c}{\textbf{KODAK}} \\
        \cmidrule(lr){1-5} \cmidrule(lr){6-7}
        \textbf{$N_G$} & \textbf{$N_{KNN}$} & \textbf{$r$} & \textbf{$\mathcal{H}_{res}$} & \textbf{$\mathcal{H}_{size}$} & \textbf{PSNR} & \textbf{MS-SSIM} \\
        \midrule
        \color{red} 250k & 16 & 0.1 & 8192 & 21 & 64.08 & 0.9999 \\
        \color{red} 100k & 16 & 0.1 & 8192 & 21 & 25.02 & 0.9442 \\
        % \color{red} 1M & 16 & 0.1 & 8192 & 21 & 85.22 & 1.0000 \\
        500k & \color{red} 8  & 0.1 & 8192 & 21 & 77.52 & 0.9999 \\
        500k & \color{red} 4  & 0.1 & 8192 & 21 & 60.06 & 0.9999 \\
        500k & 16  & 0.1 & \color{red} 4096 & 21 & 63.47 & 0.9999 \\
        500k & 16  & 0.1 & \color{red} 1024 & 21 & 41.12 & 0.9962 \\
        500k & 16  & \color{red} 0.05 & 8192 & 21 & 76.83 & 0.9999 \\
        500k & 16  & \color{red} 0.2 & 8192 & 21 & 75.94 & 0.9999 \\
        500k & 16  & 0.1 & 8192 & \color{red} 19 & 62.02 & 0.9999 \\
        500k & 16  & 0.1 & 8192 & \color{red} 17 & 55.97 & 0.9999 \\
        \textbf{500k} & \textbf{16} & \textbf{0.1} & \textbf{8192} & \textbf{21} & \textbf{77.09} & \textbf{0.9999} \\
        \bottomrule
        \end{tabular}
        \caption{
        Ablation study on the KODAK dataset.  
        We evaluate the influence of the number of Gaussian components ($N_G$), number of nearest neighbors ($N_{KNN}$), search radius ($r$), hashgrid resolution ($\mathcal{H}_{res}$), and hash map size ($\mathcal{H}_{size}$) on reconstruction quality.  
        PSNR and MS-SSIM are reported for each configuration. We marked with {\color{red}red} changed parameter.
        }
        \label{tab:ablation}
        \end{table} 

We observe that the quality of the features obtained from the hashgrid $\mathcal{H}$ has the largest impact on our model’s performance. Reducing either the hashgrid resolution $\mathcal{H}_{res}$ or the hash map size $\mathcal{H}_{size}$ results in a substantial drop in PSNR. Another important factor is the number of Gaussians $N_G$: decreasing $N_G$ by a factor of 5 leads to a threefold reduction in PSNR. Other parameters, such as neighborhood size or search radius, have a much smaller effect on reconstruction quality.

\subsection{Size and Time Comparisons}

We compare the computational characteristics of \our{} with several representative baselines, reporting training time, inference throughput, and model size. To ensure consistency across methods with differing input-size constraints, all models were trained on the DIV2K dataset \cite{Agustsson_2017_CVPR_Workshops} subjected to \(3\times\) bicubic downscaling. 
% Training was performed on an NVIDIA RTX 5090 GPU, except for MiraGe \cite{waczynskamirage}, which requires older CUDA support and was therefore trained on an NVIDIA RTX 3090. 
For \our{}, we report only the components required at inference (excluding optimizers, training buffers, and hash-grid encoders) so that the model size reflects the minimal representation used for rendering.

The results in Table~\ref{tab:size_speed} expose a clear trade-off between editability, reconstruction quality, and runtime/storage costs. Only MiraGe and GaINeR provide native editability. These two methods also achieve the highest PSNR in our experiments, indicating that editability in our setting did not come at the expense of reconstruction fidelity. However, this capability incurs a computational cost: both editable models have substantially larger serialized footprints and slower end-to-end performance compared to lightweight baselines (SIREN \cite{sitzmann2020implicit}, FINER \cite{liu2024finer}, Gauss \cite{ramasinghe2022beyond}) and extremely compact pipelines (GaussianImage \cite{zhang2024gaussianimage}). Notably, our primary editable baseline, MiraGe, is more than twice as large as our model and, due to its densification process, does not maintain a constant representation size during training, whereas our method preserves a fixed and compact footprint throughout.

\begin{table}[ht!]
    \centering
    % \resizebox{0.7\linewidth}{!}{
    \begin{tabular}{l|cc|c}
        \toprule
        \multirow{2}{*}{\textbf{Model}} & \multicolumn{2}{c|}{\textbf{Time}} & \multirow{2}{*}{\textbf{Size [MB]}} \\
        \cmidrule(lr){2-3}
         & Train [s] & Inference [FPS] & \\
        \midrule
        WIRE \cite{saragadam2023wire} & 410 & 34.01 & 0.76 \\
        SIREN \cite{sitzmann2020implicit} & 89 & 163.90 & 0.76 \\
        FINER \cite{liu2024finer} & 122 & 83.84 & 0.76 \\
        Gauss \cite{ramasinghe2022beyond} & 134 & 105.88 & 0.76 \\
        GaussianImage \cite{zhang2024gaussianimage} & 236 & 5187.82 & 0.074 \\
        MiraGe \cite{waczynskamirage} & 299 & 358.07 & 109.82 \\
        GaINeR (our) & 792 & 2.50 & 51.43 \\
        \bottomrule
    \end{tabular}
    % }
    \caption{Training time, inference speed, and model size for \our{} and baseline models. All methods were trained on the DIV2K dataset \cite{Agustsson_2017_CVPR_Workshops} downscaled \(3\times\) with bicubic filtering to satisfy input-size requirements across architectures.}
    \label{tab:size_speed}
\end{table} 

\clearpage
% \bibliography{ECCV/main}

% \end{document}

\end{document}